\documentclass[journal]{IEEEtran}

\ifCLASSINFOpdf
  \usepackage[pdftex]{graphicx}
  \graphicspath{{../pdf/}{../jpeg/}}
  \DeclareGraphicsExtensions{.pdf,.jpeg,.png}
\else
  \usepackage[dvips]{graphicx}
  \graphicspath{{../eps/}}
  \DeclareGraphicsExtensions{.eps}
\fi

\usepackage{subfigure}
\usepackage[switch]{lineno}
\usepackage{multicol}
\usepackage{multirow}
\usepackage{microtype}      
\usepackage[american]{babel}
\usepackage[ruled,linesnumbered]{algorithm2e}
\usepackage{amsmath}
\usepackage{amssymb}
\usepackage{wrapfig}
\usepackage{threeparttable}
\usepackage{flushend}
\newcommand{\eg}{\emph{e.g.}}

\newcommand{\ie}{\emph{i.e.}}

\graphicspath{{figure_compressed/}}

\begin{document}
\title{Bridge the Gap between Supervised and Unsupervised Learning for Fine-Grained Classification}

\author{Jiabao~Wang,
        Yang~Li,
        Xiu-Shen~Wei,
        Hang~Li,
        Zhuang~Miao,
        and~Rui~Zhang
\thanks{$\bullet$ This work has been supported by the Natural Science Foundation of Jiangsu Province (No. BK20200581, BK20210340), and in part by National Key R\&D Program of China (2021YFA1001100), the National Natural Science Foundation of China under Grant 61806220, the China Postdoctoral Science Foundation under Grant 2020M683754 and 2021T140799, CAAI-Huawei MindSpore Open Fund, and Beijing Academy of Artificial Intelligence (BAAI).}
\thanks{$\bullet$ Jiabao Wang, Yang Li, Hang Li, Zhuang Miao and Rui Zhang are with Army Engineering University of PLA, Nanjing 210007, China. (E-mails: jiabao\_1108@163.com; solarleeon@outlook.com; lihang0003@outlook.com; emiao\_beyond@163.com; 3959966@qq.com). Xiu-Shen Wei is with PCA Lab, Key Lab of Intelligent Perception and Systems for High-Dimensional Information of Ministry of Education, and Jiangsu Key Lab of Image and Video Understanding for Social Security, School of Computer Science and Engineering, Nanjing University of Science and Technology, Nanjing 210094, China. Xiu-Shen Wei is also with State Key Lab. for Novel Software Technology, Nanjing University, Nanjing 210023, China. (E-mail: weixs@njust.edu.cn)}
\thanks{$\bullet$ Corresponding author: Yang Li}
}

\markboth{SUBMITTED TO IEEE TIP}%
{Shell \MakeLowercase{\textit{et al.}}: Bare Demo of IEEEtran.cls for IEEE Journals}

\maketitle

\begin{abstract}
Unsupervised learning technology has caught up with or even surpassed supervised learning technology in general object classification (GOC) and person re-identification (re-ID). However, it is found that the unsupervised learning of fine-grained visual classification (FGVC) is more challenging than GOC and person re-ID. In order to bridge the gap between unsupervised and supervised learning for FGVC, we investigate the essential factors (including feature extraction, clustering, and contrastive learning) for the performance gap between supervised and unsupervised FGVC. Furthermore, we propose a simple, effective, and practical method, termed as UFCL, to alleviate the gap. Three key issues are concerned and improved: First, we introduce a robust and powerful backbone, ResNet50-IBN, which has an ability of domain adaptation when we transfer ImageNet pre-trained models to FGVC tasks. Next, we propose to introduce HDBSCAN instead of DBSCAN to do clustering, which can generate better clusters for adjacent categories with fewer hyper-parameters. Finally, we propose a weighted feature agent and its updating mechanism to do contrastive learning by using the pseudo labels with inevitable noise, which can improve the optimization process of learning the parameters of the network. The effectiveness of our UFCL is verified on CUB-200-2011, Oxford-Flowers, Oxford-Pets, Stanford-Dogs, Stanford-Cars and FGVC-Aircraft datasets. Under the unsupervised FGVC setting, we achieve state-of-the-art results, and analyze the key factors and the important parameters to provide a practical guidance.

\end{abstract}


\IEEEpeerreviewmaketitle

\section{Introduction}
\begin{figure}[!t]
\centering
\label{fig1a}
\includegraphics[width=0.49\textwidth]{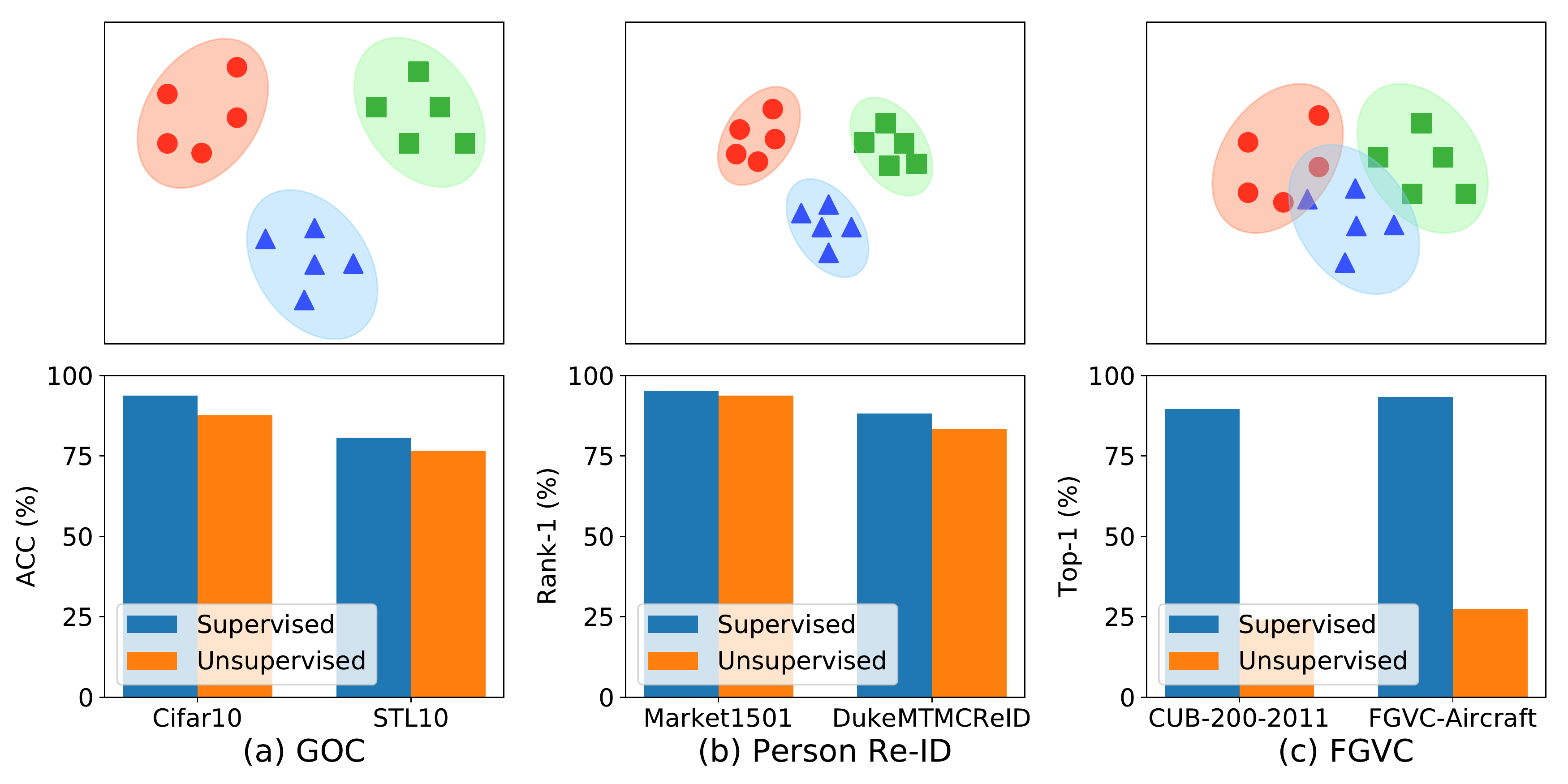}
\caption{(a) General object classification (GOC) and (b) Person re-identification (re-ID), (c) Fine-grained visual classification (FGVC). In the first row, circles, squares and triangles respectively represent three types of examples, and ellipse represents the distribution of examples from different categories, where FGVC has small difference between sub-categories and large variance within sub-categories. In the second row, CIFAR-10 and STL-10 results are from SCAN~\cite{DBLP:conf/eccv/GansbekeVGPG20}, Market-1501 and DukeMTMC-ReID results are from ICE~\cite{DBLP:journals/corr/abs-2103-16364}, CUB-200-2011 and FGVC-Aircraft results are from reproduced Group-Sampling~\cite{DBLP:journals/corr/abs-2107-03024} and PMG~\cite{DBLP:conf/eccv/DuCBXMSG20}.}
\label{fig1}
\end{figure}

\IEEEPARstart{F}ine-grained visual classification (FGVC)~\cite{DBLP:journals/corr/abs-1907-03069} is a long-standing problem in the field of computer vision, aiming to classify hundreds of subordinate categories that are under the same basic-level category, \eg, different species of birds~\cite{WahCUB_200_2011}, models of cars~\cite{DBLP:conf/iccvw/Krause0DF13}, and aircrafts~\cite{DBLP:journals/corr/MajiRKBV13}. It is a more challenging problem than general object classification (GOC) due to the inherently subtle inter-class object variations amongst sub-categories (as show in the first row of Fig.~\ref{fig1}). Besides, it is extremely hard even for human beings to recognize hundreds of sub-categories.

Early works on FGVC mostly attempt to find discriminative regions with the assistance of manual annotations. For example, Branson et al.~\cite{DBLP:journals/corr/BransonHBP14} proposed to obtain the local parts by using groups of key-points to compute multiple warped image regions. Zhang et al.~\cite{DBLP:conf/eccv/ZhangDGD14} trained a R-CNN model based on ground truth part annotations, and then performed part detection to get discriminative regions. Besides, segmentation models, such as PS-CNN~\cite{DBLP:conf/cvpr/HuangXTZ16} and Mask-CNN~\cite{DBLP:journals/pr/WeiXWS18}, were also employed to locate part regions. However, human annotations based on object parts are difficult to obtain, and can often be error-prone resulting in performance degradations.

To alleviate this burden, weakly-supervised FGVC tried to training models with only image-level labels~\cite{DBLP:conf/cvpr/FuZM17,DBLP:conf/iccv/ZhengFML17,DBLP:journals/pami/LinRM18,DBLP:conf/cvpr/WangMD18, DBLP:journals/ijcv/HePZ19,DBLP:conf/aaai/GaoHWHS20, DBLP:journals/tip/ChangDXBLMWGS20,DBLP:conf/eccv/DuCBXMSG20}. These models are able to locate more discriminative local regions for classification with less label efforts. However, label annotation of training data is also required. \emph{What happens if there is no access to ground-truth semantic labels during FGVC training? Can they achieve or reach the performance of supervised learning methods?} Unfortunately, the answer is no. As illustrated in the second row of Fig.~\ref{fig1}, the unsupervised and supervised methods have the similar performance for GOC and person re-identification (re-ID), but there is a huge gap between the unsupervised method and supervised learning methods for FGVC. This performance gap is also an undesirable consequence of slowing down the research progress in the field of unsupervised FGVC.

Because of the recent success of self-supervised learning methods in GOC~\cite{DBLP:conf/cvpr/He0WXG20,DBLP:conf/icml/ChenK0H20, DBLP:conf/nips/CaronMMGBJ20,DBLP:conf/nips/GrillSATRBDPGAP20, DBLP:conf/cvpr/ChenH21} and person re-ID~\cite{DBLP:conf/nips/Ge0C0L20,DBLP:conf/iclr/GeCL20,DBLP:journals/corr/abs-2103-16364,
DBLP:journals/corr/abs-2107-03024,DBLP:conf/cvpr/XuanZ21,DBLP:journals/corr/abs-2103-11568}, one natural question arises: \emph{What are the essential factors for the performance gap between unsupervised and supervised methods for FGVC?} Ideally, if we know the key factors of the aforementioned question and encourage a model to be robust to these factors during unsupervised learning, then we might design a more appropriate unsupervised FGVC method compared with the state-of-the-art supervised FGVC methods, which can bridge the gap between unsupervised and supervised learning for FGVC. In addition, we argue that studying this topic may also unleash the potentials of unsupervised FGVC.

To this end, in this paper, we investigate the gap difference between FGVC, GOC, and person re-ID. We argue that FGVC is more challenging due to it contains different identities with larger intra-class variances. Moreover, the performance gap between unsupervised and supervised FGVC is mainly affected by three main factors, including feature extraction, clustering, and contrastive learning, which named as \emph{Instability Gap}.

\begin{figure*}[!t]
\centering
\includegraphics[width=0.90\textwidth]{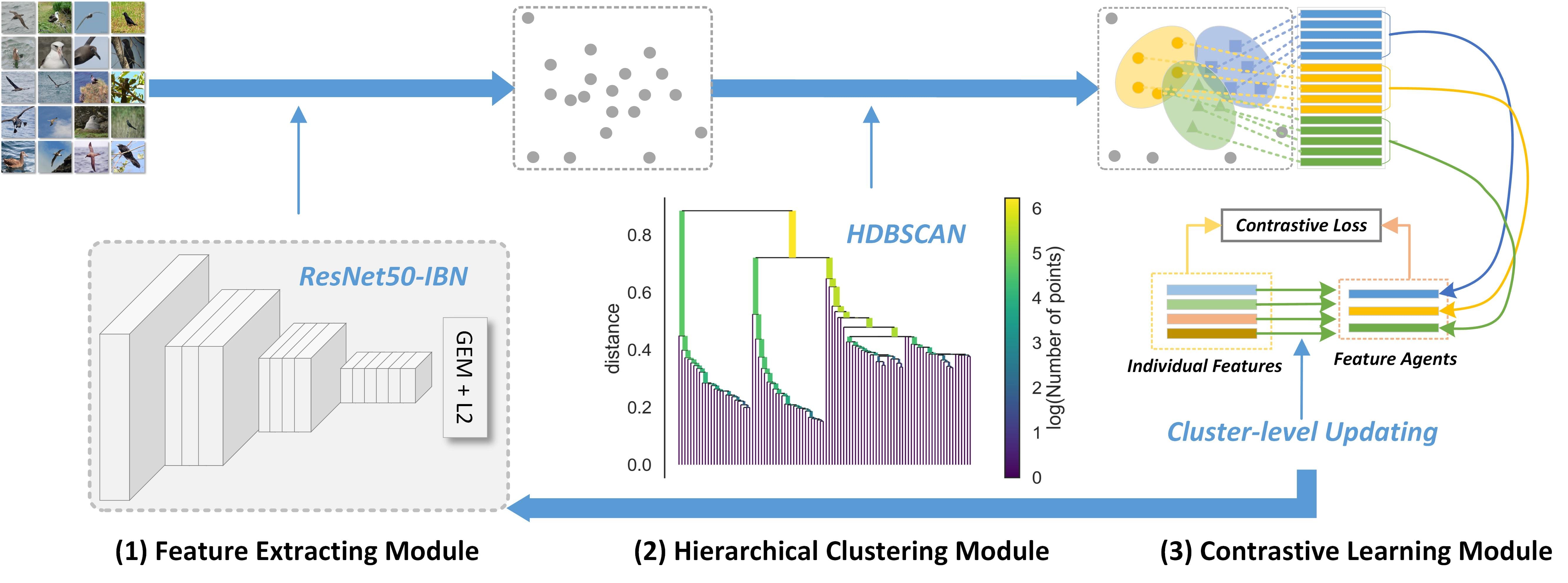}
\caption{Our UFCL includes a feature extraction module, a hierarchical clustering module and a contrastive learning module. The three modules are seamlessly integrated into a cycled framework to learning a robust model in an unsupervised schema.}
\label{fig2}
\end{figure*}

Furthermore, we propose a simple, effective, and practical method, termed as Unsupervised Fine-grained Clustering Learning (UFCL), to alleviate the \emph{Instability Gap}. UFCL involves robust feature extraction module, stable hierarchical clustering module and cluster-level contrastive learning module, so that it converges to the optimal results by the end of the training process. As shown in Fig.~\ref{fig2}, our method involves three steps: extracting features via a backbone network, generating pseudo labels by a clustering algorithm, and updating network parameters by contrastive learning. Then, the network is upgraded at the end of each epoch. The aforementioned framework brings three key issues, and we provide solutions accordingly. First, since the initial parameters of feature extraction backbone network are usually trained on large-scale general datasets, such as ImageNet and JFT300M, which has a big domain gap with unsupervised FGVC datasets, the backbone should have robust domain adaptability. As a result, we introduce a robust and powerful backbone, ResNet50-IBN~\cite{DBLP:conf/eccv/PanLST18}, which can achieve comparable improvements as domain adaptation methods when applying it to new domains. Second, as the network is upgrade with pseudo labels, it is necessary to ensure that the false pseudo labels are as few as possible. Hence, we propose to adopt the HDBSCAN~\cite{DBLP:conf/pakdd/CampelloMS13,DBLP:journals/tkdd/CampelloMZS15} algorithm to do clustering, which can generate better clusters for adjacent categories with fewer hyper-parameters than the widely used DBSCAN algorithm. Third, optimizing a deep network by pseudo labels with inevitable noise may also cause unstable gradients, which degenerates the learning ability of unsupervised FGVC. Consequently, we introduce cluster-level contrastive leaning and propose a new weighted updating strategy to compute a feature agent for each category in a batch and update the parameters with the robust gradients of feature agents.

The effectiveness of our UFCL method is verified on six unsupervised FGVC tasks on the standard vision setting, we achieve the state-of-the-art performance with 69.0\%, 90.1\%, 79.0\%, 58.5\%, 33.7\%, and 43.3\% classification Top-1 accuracy on CUB-200-2011, Oxford-Flowers, Oxford-Pets, Stanford-Dogs, Stanford-Cars, and FGVC-Aircraft, respectively. In addition, we demonstrate the benefits of robust feature extraction module, stable hierarchical clustering module and cluster-level contrastive learning module on CUB-200-2011 and Oxford-Flowers datasets.
Comparing with the widely used ResNet50 backbone, ResNet50-IBN has 24.1\% and 1.9\% improvements on Top-1 accuracy. HDBSCAN increases the Top-1 accuracy of 3.7\% and 0.5\% compared with DBSCAN~\cite{DBLP:conf/kdd/EsterKSX96}, and the best weighted updating strategy can improve the Top-1 accuracy over 22.0\% and 11.0\% comparing with Cluster-Contrast~\cite{DBLP:journals/corr/abs-2103-11568}.

The contributions of this paper are concluded as follows:

\begin{itemize}
\item We investigate the essential factors (\ie, feature extraction, clustering, and contrastive learning) for the performance gap between unsupervised and supervised FGVC, and we propose a simple, effective, and practical method, termed as UFCL, to alleviate the gap.
\item Three key issues are concerned and improved. We introduce a robust and powerful backbone, ResNet50-IBN~\cite{DBLP:conf/eccv/PanLST18}, which has an ability of domain adaptation when we transferred the ImageNet pre-trained backbone to FGVC task. Next, we propose to introduce HDBSCAN instead of DBSCAN to do clustering, which can generate better clusters for adjacent categories with fewer hyper-parameters. Furthermore, we propose a new weighted updating strategy to update the network supervised by the pseudo labels with inevitable noise, which can obtain robust gradients and improve the optimization process of parameter learning.
\item The effectiveness of our UFCL method is verified on six FGVC datasets. Under the unsupervised FGVC tasks setting, we achieve the state-of-the-art results and analyze the key factors and the important parameters to provide a practical guidance.
\end{itemize}

\section{Related Works}
Our UFCL belongs to unsupervised learning, which has aroused extensive interests due to its ability of enabling models to be trained by unlabeled data and save expensive annotation cost. The kernel idea of our UFCL comes from the combination of advantages of clustering~\cite{DBLP:conf/eccv/CaronBJD18,DBLP:conf/nips/Ge0C0L20} and self-supervised contrastive learning~\cite{DBLP:journals/corr/abs-2008-11702,DBLP:journals/corr/abs-2103-11568}.

\subsection{Clustering}
Clustering methods~\cite{DBLP:conf/eccv/CaronBJD18,DBLP:conf/nips/CaronMMGBJ20,DBLP:conf/nips/Ge0C0L20} employ cluster indexes as pseudo labels to train the model by unsupervised learning paradigm. There are two types of success applications: the pre-trained model for general object representation~\cite{DBLP:conf/eccv/CaronBJD18,DBLP:conf/nips/CaronMMGBJ20} and the special model for unsupervised person re-ID~\cite{DBLP:conf/nips/Ge0C0L20,DBLP:journals/corr/abs-2103-11568}.

In general object representation, DeepCluster~\cite{DBLP:conf/eccv/CaronBJD18} is the first unsupervised learning method that alternates between clustering the representation by \emph{k}-means and learning to predict the cluster assignments as pseudo-labels. It can scale to large-scale dataset and can be used for pre-training of supervised networks~\cite{DBLP:conf/iccv/CaronBMJ19}. Another famous one is SwAV~\cite{DBLP:conf/nips/CaronMMGBJ20}, which incorporates clustering into a Siamese network, by computing the assignment from one view and predicting it from another view. The results of these methods are affected by the clustering algorithm, which may generate false pseudo labels for unlabeled data and supervise network learning in a bad way.

In unsupervised person re-ID, the early works are dominated by unsupervised domain adaptive learning methods, include MMT~\cite{DBLP:conf/iclr/GeCL20} and SpCL~\cite{DBLP:conf/nips/Ge0C0L20}. They firstly adopted a pre-trained ImageNet model to fine-tune it on a large-scale supervised person re-ID dataset, and then to fine-tune it on the small-scale unsupervised target dataset. They distinguished trusted examples from untrusted examples through DBSCAN clustering, which has no prior requirements for data distribution and the total number of clusters as \emph{k}-means. However, these methods relied on labeled large-scale source domain datasets, and it is difficult and costs a lot to build such datasets in practice. Recent works directly conducted unsupervised clustering learning on the person re-ID datasets. For example, Group-Sampling~\cite{DBLP:journals/corr/abs-2107-03024} carried out sampling learning in the unit of group examples, which can cluster similar examples better. Cluster-Contrast~\cite{DBLP:journals/corr/abs-2103-11568} directly updated the class center on the target domain dataset, and looked for difficult examples in batch to improve the learning effect. IICS~\cite{DBLP:conf/cvpr/XuanZ21} and ICE~\cite{DBLP:journals/corr/abs-2103-16364} used the a priori distributed within the camera to further improve the performance.

The aforementioned two types of clustering methods use \emph{k}-means and DBSCAN clustering algorithms to generate pseudo labels respectively. Although DBSCAN has stronger density clustering ability and better anti-noise ability than \emph{k}-means, it is easy to cluster two adjacent categories into one cluster in the face of FGVC tasks with small inter-class difference and large intra-class variance. To alleviate the problem of DBSCAN, we propose to introduce HDBSCAN~\cite{DBLP:conf/pakdd/CampelloMS13,DBLP:journals/tkdd/CampelloMZS15} into FGVC tasks to better cluster adjacent sub-categories.

\subsection{Self-supervised Contrastive Learning}
Contrastive learning methods~\cite{DBLP:conf/icml/ChenK0H20, DBLP:conf/cvpr/He0WXG20, DBLP:conf/nips/GrillSATRBDPGAP20, DBLP:conf/cvpr/ChenH21} currently achieve state-of-the-art performance in self-supervised learning. Contrastive approaches learn the discriminative representation by bringing representation of different views of the same image closer, and spreading representations of views from different images apart~\cite{DBLP:conf/cvpr/WuXYL18}. The existing methods can be categorized into instance-level methods~\cite{DBLP:conf/icml/ChenK0H20, DBLP:conf/cvpr/He0WXG20, DBLP:conf/cvpr/ChenH21} and cluster-level methods~\cite{DBLP:journals/corr/abs-2008-11702,DBLP:conf/nips/CaronMMGBJ20,DBLP:journals/corr/abs-2103-11568}.

Instance-level methods treated views of the same image as positive pairs and views of different images in the same batch (or memory bank) as negative pairs. For example, SimCLR~\cite{DBLP:conf/icml/ChenK0H20} treated the examples in the batch as the negative examples. MoCo~\cite{DBLP:conf/cvpr/He0WXG20} adopted a dictionary to implement contrastive learning, where one branch of the Siamese network is updated with momentum strategy. SiaSiam~\cite{DBLP:conf/cvpr/ChenH21} trained the Siamese network by stopping gradient back-propagation in one of the branch. However, instance-level methods simply make each example independent and repel to each other, which ignores the cluster structure information in examples.

Cluster-level methods regard examples in the same clusters as positive cluster and other examples as negative clusters. InterCLR~\cite{DBLP:journals/corr/abs-2008-11702} used both InfoNCE loss and MarginNCE loss to attract positive examples and repelled negative examples. SwAV~\cite{DBLP:conf/nips/CaronMMGBJ20} proposed an online clustering loss to improve the ability of the network to explore the inter-invariance of clusters. Cluster-Contrast~\cite{DBLP:journals/corr/abs-2103-11568} adopted ClusterNCE loss to compute the batch-hard example with the cluster centers in memory bank, which greatly improve the performance in person re-ID. However, cluster-level methods rely on the clustering results in which the batch-hard example learning faces convergence problem in the early training stage.

The aforementioned methods carry out self-supervised contrastive learning from different granularity. The cluster-level methods use the structure information between different identities and are more robust than the instance-level ones, but it is also a challenge problem for FGVC to solve the convergence in the early training stage caused by the batch-hard example learning strategy. In this paper, based on cluster-level contrastive learning, we propose a weighted feature agent, and robustly updated it through a new centroid-based updating mechanism.

\subsection{Unsupervised Fine-grained Classification}
FGVC faces greater challenges due to the subtle difference among sub-categories and large variations of many different identities of the same sub-categories. At present, it is dominated by supervised learning methods, such as PA-CNN~\cite{DBLP:journals/tip/ZhengFZLM20}, Cross-X~\cite{DBLP:conf/iccv/LuoYML0LYL19}, PMG~\cite{DBLP:conf/eccv/DuCBXMSG20}, CAMF~\cite{DBLP:journals/spl/MiaoZWLL21}. The existing recent methods~\cite{DBLP:conf/iccv/LuoYML0LYL19,DBLP:journals/spl/MiaoZWLL21} focused on the attention mechanism to find effective object discriminative parts to improve the effectiveness. However, the small inter-class difference is hard to identify even for professional experts, and annotating these fine-grained sub-categories requires a lot of labor costs.

Therefore, researchers try to explore the unsupervised fine-grained image classification. However, the performance of existing unsupervised methods reported on fine-grained classification datasets is very low. For example, the cluster method, AND~\cite{DBLP:conf/icml/HuangDGZ19}, only reaches 14.4\% and 32.3\% on CUB-200-2011 and Stanford-Dogs datasets respectively.
Although it enables the control and mitigation of the clustering errors and their negative propagation, it cannot tackle the non-gaussian distribution and the inevitable noise. The unsupervised contrastive learning methods, SimCLR~\cite{DBLP:conf/icml/ChenK0H20} and BYOL~\cite{DBLP:conf/nips/GrillSATRBDPGAP20}, evaluate the fine-grained classification tasks, but the final linear classifiers needs to be trained with labels. In this paper, we directly reported the accuracy based on the $k$-NN classifier, which does not require training. Besides, based on the priority of unsupervised clustering and contrastive learning, in this paper, we incorporate both of them into one joint framework to iteratively generate pseudo labels and train the model, in a robust schema.

\section{Our Approach}
\label{sec:approach}
The framework of our proposed UFCL is shown in Fig.~\ref{fig2}. It is a learning framework which contains three key modules. The first one is feature extraction module, which is a deep neural network to extract features directly from images. The second one is hierarchical clustering module, which is a robust algorithm to cluster the features to assign pseudo labels to the images. And the third one is contrastive learning module, which is an updating strategy to learn the parameters of the deep neural network by a cluster-level contrastive loss. All the three modules are seamlessly integrated into a cycled framework to learning a robust model in an unsupervised schema. Hence, different modules can give feedback to each other in UFCL, which results in bridging the gap between supervised and unsupervised learning for FGVC.

Ideally, a robust feature extraction module can generate effective features, which can be clustered into stable clusters through a stable clustering algorithm. Pseudo labels with high accuracy can be generated from the stable clusters, and with the help of the contrastive learning module, the parameters of deep neural network in feature extraction module can be updated effectively. As a result, it is a key task to systematically analyze and design these three modules. In the following subsections, we will focus on why and how to design three key modules to bridge the gap between supervised and unsupervised learning for FGVC, and present the details of our UFCL framework.

\subsection{Feature Extraction Module}
\label{sec:featureExtraction}
Our UFCL contains a feature extraction backbone network from~\cite{DBLP:conf/eccv/PanLST18} as a module, which is used to extract robust the feature representation for given images. More specifically, ResNet50-IBN~\cite{DBLP:conf/eccv/PanLST18} is constructed based on ResNet-50 by replacing batch normalization block with IBN block, which is composed of Batch Normalization (BN) and Instance Normalization (IN), to extract robust features. BN can can accelerate training and preserves discriminative features, while IN can provide visual and appearance invariance. The built-in appearance invariance introduced by IN helps the model to generalize from the pre-trained ImageNet domain to the FGVC domain, even without using the data from the target domain~\cite{DBLP:conf/eccv/PanLST18}.

Of course, the backbone is not limited to ResNet50-IBN, other CNN-based~\cite{DBLP:journals/corr/abs-1910-06827} or Transformer-based~\cite{DBLP:journals/corr/abs-2109-06165} backbones with domain adaptation ability, can also be used to substitute the ResNet50-IBN as our backbone. But it is not the focus of this paper to study new backbone networks. Hence, we just adopted ResNet50-IBN for illustrating the effectiveness of our UFCL framework, and leave the study of other candidate networks for future pursuit.

After all convolutional layers of ResNet50-IBN, a generalized pooling layer, GEM pooling~\cite{DBLP:journals/pami/RadenovicTC19}, is introduced to offer significant performance boost over standard non-trainable pooling layer. Specifically, suppose that a feature tensor $\mathbf{X} \in {\mathbb{R}^{W\times H\times K}}$ extracted for a given image $I$ by the ResNet50-IBN backbone, where $W$, $H$, and $K$ are the width, height and number of channels of $\mathbf{X}$ respectively. The feature matrix of the $k$-th channel of $\mathbf{X}$ is expressed as $\mathbf{X}(k)\in {{R}^{W \times H}}$, and then a real-value $f(k)$ is generated by doing GEM on $\mathbf{X}(k)$:

\begin{equation}\label{eq1}
\begin{aligned}
f(k)={{\left( \frac{1}{|\mathbf{X}(k)|}\sum\limits_{x\in \mathbf{X}(k)}{{{x}^{p(k)}}} \right)}^{\frac{1}{p(k)}}},
\end{aligned}
\end{equation}
where $p(k)$ represents the super-parameter corresponding to the $k$-th channel, and can be learned in the training process. When $p(k)\to \infty $, it equivalents to GMP. When $p(k)=1$, it equivalents to GAP. GEM pooling generalizes max and average pooling to improve the feature presentation for the next clustering step. Finally, the feature vector $\mathbf{f}$ is obtained by doing the $\ell_2$ regularization on ${{\mathbf{f}}^{gem}}= [f(1),f(2),...,f(K)]$.

\subsection{Hierarchical Clustering Module}
\label{sec:hierarchicalClustering}
Based on the features extracted from the feature extraction module, a hierarchical clustering algorithm, HDBSCAN~\cite{DBLP:conf/pakdd/CampelloMS13,DBLP:journals/tkdd/CampelloMZS15}, is introduced to cluster the features into many clusters. It generates a pseudo label for the feature of an image according to the cluster to which it belongs.

\begin{figure}[!t]
\centering
\subfigure[Results of DBSCAN]{
\label{fig3a}
\includegraphics[width=0.23\textwidth]{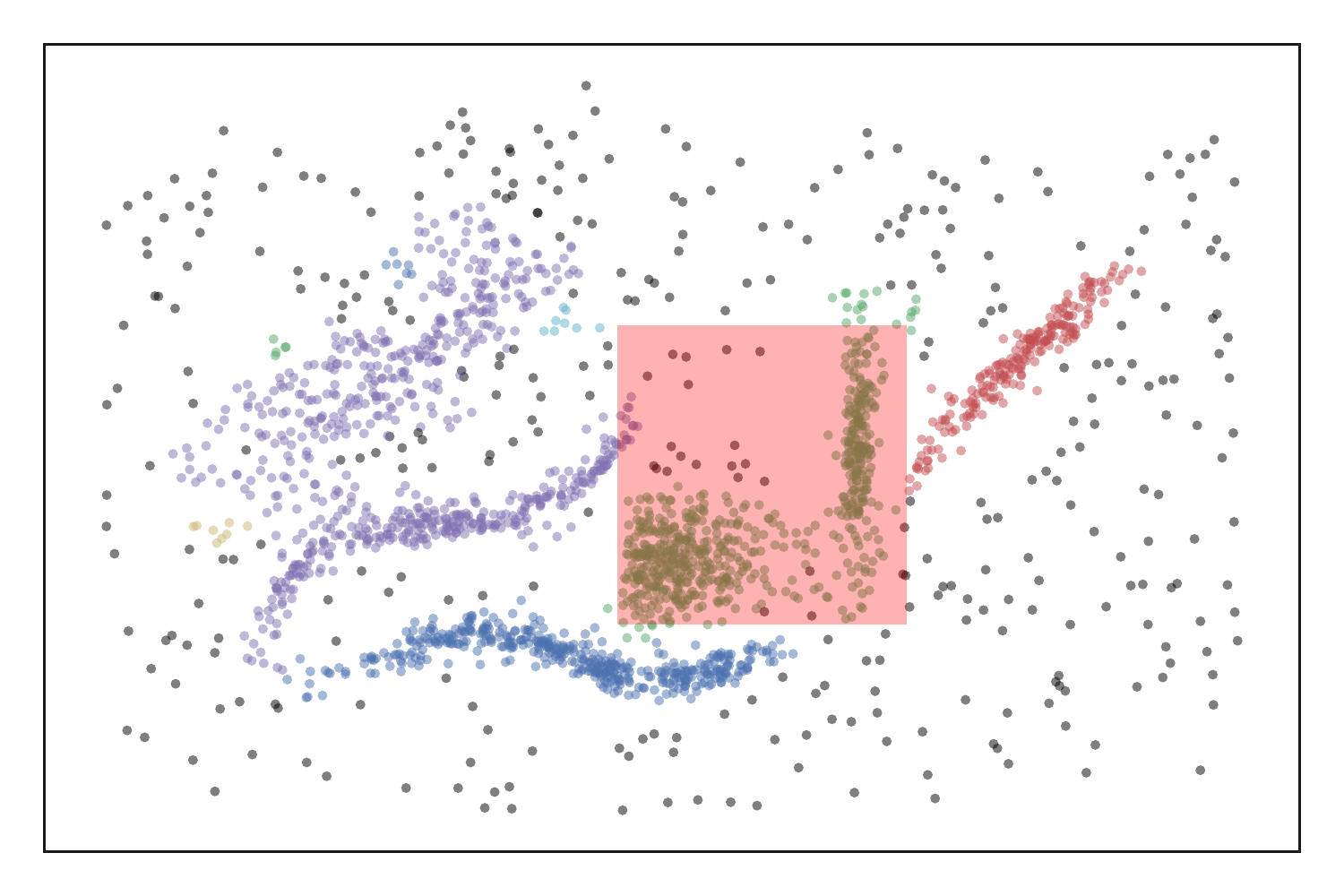}}
\subfigure[Results of HDBSCAN]{
\label{fig3b}
\includegraphics[width=0.23\textwidth]{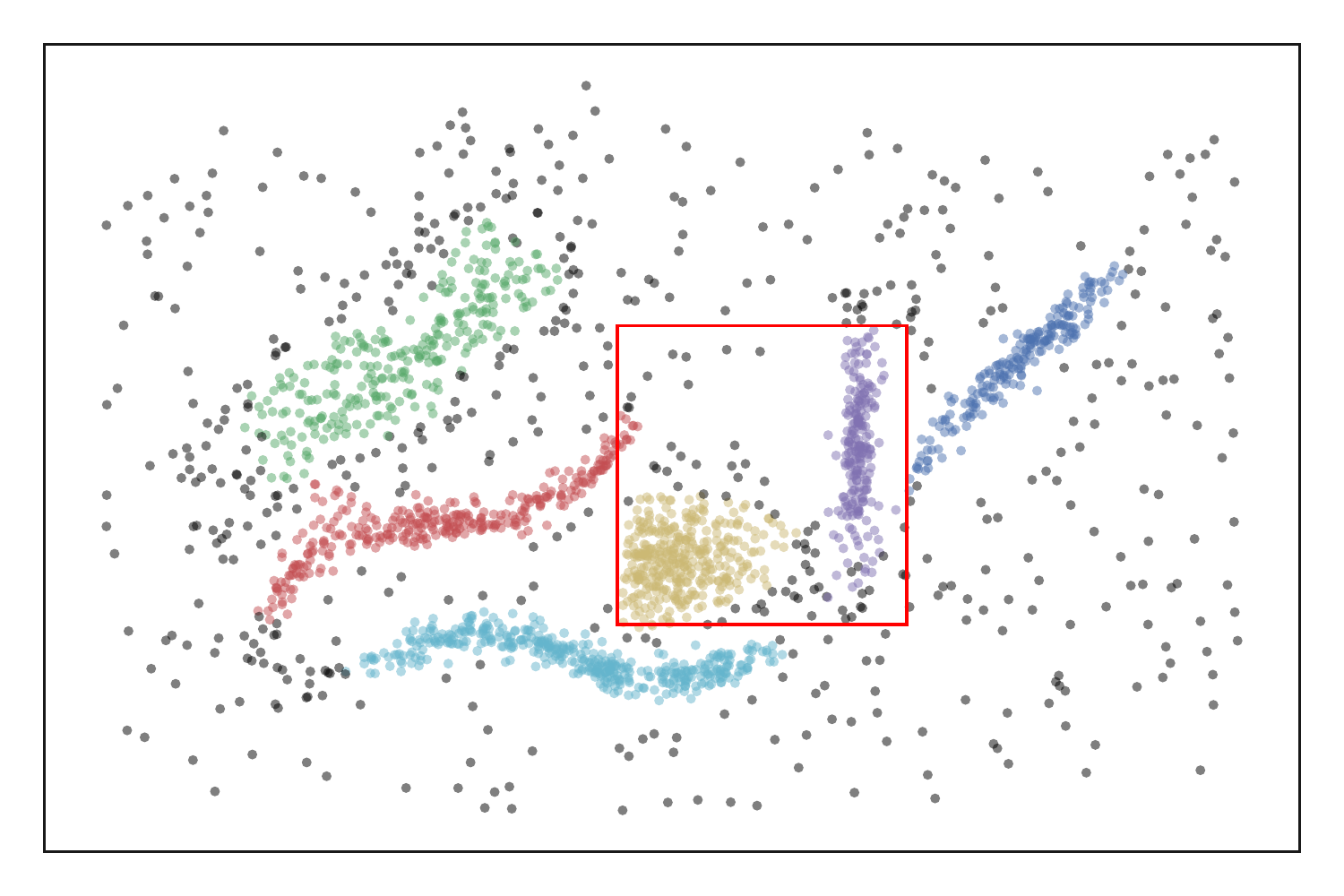}}
\caption{(a) Two adjacent categories are false clustered together (red region), (b) The two adjacent categories are spited as different classes correctly~\cite{DBLP:journals/tkdd/CampelloMZS15} (red box).
}
\label{fig3}
\end{figure}

As we known, the density-based clustering algorithm, DBSCAN~\cite{DBLP:conf/kdd/EsterKSX96}, is widely used for unsupervised learning in person re-ID~\cite{DBLP:journals/corr/abs-2107-03024, DBLP:conf/nips/Ge0C0L20,DBLP:journals/corr/abs-2108-03439, DBLP:journals/corr/abs-2109-12333, DBLP:journals/corr/abs-2103-16364, DBLP:journals/corr/abs-2109-14157, DBLP:journals/corr/abs-2103-11568}. DBSCAN can automatically produce a number of categories based on the distribution of the examples, and has no prior requirements for data distribution and the total number of clusters. It takes two parameters, $\varepsilon $ and $k$, where $\varepsilon $ represents a distance scale, and $k$ is a density threshold expressed in terms of a minimum number of points. However, the fixed distance scale $\varepsilon $ faces the problem of variable density clustering, which DBSCAN struggles with. As shown in Fig.~\ref{fig3a}, for adjacent categories with multiple adjacent examples, DBSCAN force examples of two adjacent categories into one cluster (red region in figure) if the $\varepsilon $ is set to a single fixed value. In FGVC tasks, the inter-class difference are small, features of the examples from adjacent categories are close to each other, and they is easy to be clustered into one cluster when DBSCAN is adopted.

To solve the aforementioned problem, we propose to introduce HDBSCAN~\cite{DBLP:conf/pakdd/CampelloMS13,DBLP:journals/tkdd/CampelloMZS15} to address variable density clustering problem. HDBSCAN extends DBSCAN by converting it into a hierarchical clustering algorithm and extracts the stable and flat clusterings from a condensed tree. HDBSCAN builds a hierarchy of DBSCAN for varying $\varepsilon $ values, and encourages the algorithm to search the best parameter over all $\varepsilon $ values, so it can deal with variable density clustering problem. As shown in Fig.~\ref{fig3b}, HDBSCAN can split two adjacent categories in two clusters (red box in figure) by introducing a notion of \emph{minimum cluster size}, which can find the clusters that persist for many values of $\varepsilon $. Besides, HDBSCAN has only one parameter, \emph{minimum cluster size}, to be set, so it is more easy to find the best parameter than DBSCAN.

In our implementation, HDBSCAN uses the pre-computed Jaccard distance to do clustering based on the features of all training examples, and generates a large number of clusters. For clustering results, the examples belonging to the same cluster are classified into a same class and generate the same pseudo label. The pseudo labels of examples among different clusters are different. The examples not belonging to any cluster are regarded as outliers, and pseudo labels are not assigned for them and they will not take part in the subsequent training process.

\subsection{Contrastive Learning Module}
\label{sec:contrastiveLearning}
The goal of contrastive learning module is to learn the parameters of the backbone network in contrastive fashion with the assigned pseudo labels. It relies on the images and their generated pseudo labels to optimize the parameters and make them more suitable for FGVC tasks. State-of-the-art unsupervised learning methods in person re-ID compute the contrastive loss between query instances and memory dictionary. MMCL~\cite{DBLP:conf/cvpr/WangZ20a} computes the loss and updates the memory bank both in the instance level. SpCL~\cite{DBLP:conf/nips/Ge0C0L20} and Group-Sampling~\cite{DBLP:journals/corr/abs-2107-03024} computes the loss in cluster level but updates the memory dictionary in the instance level. Due to the number of instances in each cluster is imbalanced, the updating progress for each cluster is inconsistent correspondingly. To solve the inconsistency problem, ClusterContrast~\cite{DBLP:journals/corr/abs-2103-11568} computes the loss and update the dictionary both in the cluster level. It uses the feature of a random instance in the cluster to initialize the cluster feature, and select the hardest instance for each person identity in the mini-batch to momentum update the cluster feature. However, it is inevitable to contain noisy labels in the pseudo labels, so the random instance initialization and hardest instance updating mechanism are easy to mislead the network optimization.

To solve aforementioned problem, we propose a weighted feature agent for the cluster-level memory bank, and propose a centroid-based updating mechanism to update the feature agent. The feature agent, calculated by weighting all examples coming from the same cluster and momentum updated based on the mean feature of the same cluster in the mini-batch, is robust to the noisy examples, especially for the unsupervised FGVC that relies heavily on the accuracy of the pseudo labels to distinguish two categories with subtle variance. It can help feature extraction network learning parameters more effectively.

In the training process, the features are extracted by forward propagation, and the ClusterNCE loss~\cite{DBLP:journals/corr/abs-2103-11568} is calculated based on the mean feature of each class in mini-batch and the feature agent in cluster-level memory bank. The ClusterNCE loss function is:

\begin{equation}\label{eq7}
\begin{aligned}
{{L}}=-\log \frac{\exp (d({{\mathbf{\bar{f}}}_{k}},{{\mathbf{c}}_{k}})/\tau )}{\sum\nolimits_{j}{\exp (d(\mathbf{m_j},{{\mathbf{c}}_{j}})/\tau )}},
\end{aligned}
\end{equation}
where ${{\mathbf{\bar{f}}}_{k}}$ denotes the mean feature of the $k$-th class in each mini-batch, ${{\mathbf{c}}_{k}}$ represents the feature agent of $k$-th cluster, and $\tau $ represents the temperature super-parameter. $d({{\mathbf{\bar{f}}}_{k}},\mathbf{c}_k)$ represents the Euclidean distance between the mean feature ${{\mathbf{\bar{f}}}_{k}}$ and the feature agent $\mathbf{c}_k$ of the $k$-th class. When ${{\mathbf{\bar{f}}}_{k}}$ has a higher similarity to its ground-truth feature agent ${{\mathbf{c}}_{k}}$ and dissimilarity to all other feature agents, the objective loss has a lower value.

In the following subsections, we discuss about the computation and updating of feature agent in detail.

\subsubsection{Computing Feature Agent}
After generating pseudo labels, a feature agent for a cluster is calculated. The calculation method is obtained by the following weighted averaging:

\begin{equation}\label{eq2}
\begin{aligned}
{{\mathbf{c}}_{k}}\text{=}\sum\limits_{i=1}^{{{N}_{k}}}{w_{k}^{i}\mathbf{f}_{k}^{i}}.
\end{aligned}
\end{equation}

In Eq. (\ref{eq2}), ${{\mathbf{c}}_{k}}$ is the feature agent (weighted centroid) of $k$-th cluster, ${{N}_{k}}$ is the number of examples in $k$-th cluster, $\mathbf{f}_{k}^{i}$ is the feature of $i$-th example of $k$-th cluster, and $w_{k}^{i}$ is the corresponding weight, which is obtained by the following formula:

\begin{equation}\label{eq3}
\begin{aligned}
w_{k}^{i}=\frac{\exp (d(\mathbf{f}_{k}^{i},{{\mathbb{F}}_{k}}))}{\sum\nolimits_{j}{\exp (d(\mathbf{f}_{k}^{j},{{\mathbb{F}}_{k}}))}},
\end{aligned}
\end{equation}

In Eq. (\ref{eq3}), ${{\mathbb{F}}_{k}}$ is the set of all example features of $k$-th cluster, $\exp(\cdot)$ represents the exponential function, and the distance measurement $d(\mathbf{f}_{k}^{i},{{\mathbb{F}}_{k}})$ can be calculated by any of the following three calculation methods:

\textbf{(a) zero distance:}
\begin{equation}\label{eq4}
\begin{aligned}
d(\mathbf{f}_{k}^{i},{{\mathbb{F}}_{k}})\text{=}0,
\end{aligned}
\end{equation}
where the distance constraint is out of consideration, which is called \emph{zero distance}. This distance makes the weight $w_{k}^{i}=1/{{N}_{k}}$.

\textbf{(b) min distance:}
\begin{equation}\label{eq5}
\begin{aligned}
d(\mathbf{f}_{k}^{i},{{\mathbb{F}}_{k}})=\arg \underset{\mathbf{f}_{k}^{j}\in {{\mathbb{F}}_{k}},j\ne i}{\mathop{\min }}\,\{d(\mathbf{f}_{k}^{i},\mathbf{f}_{k}^{j})\},
\end{aligned}
\end{equation}
where $d(\mathbf{f}_{k}^{i},{{\mathbb{F}}_{k}})$ is the minimum distance from the feature $\mathbf{f}_{k}^{i}$ to all other example features of $k$-th class is called \emph{min distance}.

\textbf{(c) mean distance:}
\begin{equation}\label{eq6}
\begin{aligned}
d(\mathbf{f}_{k}^{i},{{\mathbb{F}}_{k}})=\frac{1}{{{N}_{k}}}\sum\limits_{j=0,j\ne i}^{{{N}_{k}}}{d(\mathbf{f}_{k}^{i},\mathbf{f}_{k}^{j})},
\end{aligned}
\end{equation}
where $d(\mathbf{f}_{k}^{i},{{\mathbb{F}}_{k}})$ represents the average distance from the feature $\mathbf{f}_{k}^{i}$ to all other example features of $k$-th class, which is called \emph{mean distance}.

\subsubsection{Updating Feature Agent}
The feature agent is momentum updated with Eq. (\ref{eq8}):

\begin{equation}\label{eq8}
\begin{aligned}
{{\mathbf{c}}_{k}}\leftarrow m{{\mathbf{c}}_{k}}+(1-m){{\mathbf{\bar{f}}}_{k}} ,
\end{aligned}
\end{equation}
where $m$ represents the updating momentum, and ${{\mathbf{\bar{f}}}_{k}}$ is calculated as follows:

\begin{equation}\label{eq9}
\begin{aligned}
{{\mathbf{\bar{f}}}_{k}} = \frac{1}{{|{\mathbb{B}}_{k}|}}\sum\nolimits_{\mathbf{f}\in {{\mathbb{B}}_{k}}}{\mathbf{f}}\,,
\end{aligned}
\end{equation}
where ${\mathbb{B}}_{k}$ represents the set of features belonging to the $k$-th class in a mini-batch batch, $|{\mathbb{B}}_{k}|$ represents the number of feature in set ${\mathbb{B}}_{k}$, ${\mathbf{f}}$ is an instance feature. Eq. (\ref{eq8}) represents that the feature agent uses the mean feature of each class in mini-batch for momentum updating.
Even if there is a small amount of noise examples in one cluster, the updating can avoid the influence of the noise examples when we use the feature agent, which is initialized by weighting all examples coming from the same cluster and is momentum updated based on the mean feature of the same cluster in the mini-batch. 

\section{Experiments}
\subsection{Datasets, Parameter Setting and Evaluation Criteria}

\subsubsection{Datasets} We use six widely used fine-grained classification datasets, including CUB-200-2011\footnote{http://www.vision.caltech.edu/visipedia/CUB-200-2011.html}, Oxford-Flowers\footnote{https://www.robots.ox.ac.uk/\~vgg/data/flowers/102/}, Oxford-Pets\footnote{https://www.robots.ox.ac.uk/\~vgg/data/pets/}, Stanford-Dogs\footnote{http://vision.stanford.edu/aditya86/ImageNetDogs/},
Stanford Cars\footnote{https://ai.stanford.edu/\~jkrause/cars/car\_dataset.html},
FGVC Aircraft\footnote{https://www.robots.ox.ac.uk/\~vgg/data/fgvc-aircraft/}. The statistical information on the training and testing of the six datasets is shown in Table \ref{tab1}.

\begin{table}[!t]
\centering
\caption{Detailed statistics of the datasets. `\#Class': the number of classes, `\#Train': the number of training examples, `\#Test': the number of testing examples, and `\#Images': the total number of examples.}
\label{tab1}
\begin{tabular}{|c|r|r|r|r|}
\hline
Datasets        & \#Class & \#Train & \#Test & \#Images \\ \hline
CUB-200-2011    & 200     & 5994    & 5794   & 11788    \\
Oxford-Flowers	& 102     & 2040	& 6149	 & 8189     \\
Oxford-Pets	    & 37      & 3680	& 3669	 & 7349     \\
Stanford-Dogs	& 120	  & 12000	& 8580	 & 20580    \\
Stanford Cars   & 196     & 8144    & 8041   & 16158    \\
FGVC Aircraft   & 100     & 6667    & 3333   & 10000    \\ \hline
\end{tabular}
\end{table}

\subsubsection{Parameter Setting} In implementation, the convolution stride of the first layer in the fourth stage of ResNet50-IBN is adjusted to 1, that is, the size of the feature map is not down-sampled. For the output features at the end of the fourth stage, GEM pooling~\cite{DBLP:journals/pami/RadenovicTC19} is used to merge the features of each channel to obtain the features with a dimension of 2048. Finally, $\ell_2$ normalization is performed to obtain the normalized features for clustering.

HDBSCAN is used for clustering, and the parameter \emph{min-cluster-size} is uniformly set to 5. The input image is uniformly scaled to 224$\times$224, using data augmentation such as random clipping, horizontal flipping and random erasing. The batch-size is set to 256 unless special statement. The backbone network model adopts ResNet50-IBN, initialized with ImageNet pre-trained parameters, and the parameters are updated with Adam optimizer. The initial learning rate is 0.00035 and weight decay=$5\times {10}^{-4}$, momentum update parameter $m=0.1$, and the total number of iterations is 50.

To compare different unsupervised methods for fine-grained classification, we reproduce Group-Sampling, Cluster-Contract and ICE and transfer them from person re-ID to FGVC task. The reproduction effects of each method on different datasets are shown in Table 3. When Group-Sampling is reproduced, the parameter \emph{n-group} is set to 256, and all other parameters adopt the default parameters. Cluster-Contract adopts default parameters. ICE sets batch-size to 64, and other are default parameters. On Oxford-Flowers and Oxford-Pets datasets, the number of iterations within each epoch is set to 25 and the other datasets are set to 100.

\subsubsection{Evaluation Criteria} the widely used evaluation criteria for unsupervised clustering are ACC, NMI and ARI~\cite{DBLP:conf/eccv/GansbekeVGPG20}, but these criteria are mainly used for \emph{k}-means, that is, all examples have predicted labels. Different from \emph{k}-means, DBSCAN does not predict label for all examples, so its criteria needs to be modified.

Suppose that $N$ represents the total number of all examples and $\Omega =\{{{m}_{1}},...,{{m}_{k}},...,{{m}_{K}}\}$ represents the division of class clusters, where the number of examples in the $k$-th class cluster ${{m}_{k}}$ is $\#{{m}_{k}}$, then $\sum\nolimits_{k=1}^{K}{\#{{m}_{k}}}=M, M\le N$. $C=\{{{c}_{1}},...,{{c}_{j}},...,{{c}_{J}}\}$ indicates the truth division, where the number of examples in the $j$-th class cluster ${{c}_{j}}$ is $\#{{c}_{j}}$, then $\sum\nolimits_{j=1}^{J}{\#{{c}_{j}}}=N$. Among them, the clustering results of HDBSCAN or DBSCAN clustering algorithm will have many outliers, which do not belong to any cluster, so the modified indicators are as follows:

\textbf{(a) Clustering Criteria (ACC)}

\begin{equation}\label{eq10}
\begin{aligned}
\text{ACC}=\frac{1}{N}\sum\limits_{i=1}^{M}{\delta ({{g}_{i}},\rho ({{p}_{i}}))},
\end{aligned}
\end{equation}
where ${{g}_{i}}$ is the truth label of the $i$-th example in the clustering, ${{p}_{i}}$ is the pseudo label corresponding to its prediction, and $\rho (\cdot )$ is the redistribution mapping function of the best class label, which can be realized in polynomial time by Hungarian algorithm. $\delta (a,b)$ is Dirac delta function which returns 1 if $a=b$, and 0 otherwise. Similarly, NMI and ARI are calculated in the similar way.

\textbf{(b) Classification Criteria (Top-1)}

Weighted $k$-NN classifier ($k=5$) in
AND~\cite{DBLP:conf/eccv/GansbekeVGPG20} is used for prediction,

\begin{equation}\label{eq11}
\begin{aligned}
\text{Top-1}=\frac{1}{N}\sum\limits_{c=1}^{N}{\delta ({{g}_{c}},{{p}_{c}})},
\end{aligned}
\end{equation}
where ${{g}_{c}}$ is the truth label of the $c$-th example in the clustering result and ${{p}_{c}}$ is the predicted pseudo label. The calculation method is

\begin{equation}\label{eq12}
\begin{aligned}
{{p}_{c}}=\sum\nolimits_{i\in {{N}_{k}}}{{{w}_{i}}\cdot \delta (c,{{c}_{i}})},
\end{aligned}
\end{equation}
where $\delta (c,{{c}_{i}})$ means to judge whether the $c$-th example and the $c_i$-th example are k-nearest neighbors. ${{w}_{i}}=\exp ({{p}_{i}}/\tau )$, $\tau =0.07$.

\subsection{Results and Analysis}

\subsubsection{Basic Results and Analysis}
At present, there is few reports on the unsupervised learning for fine-grained classification. Therefore, we directly transfer the existing unsupervised learning method from the domain of person re-ID, and mainly modify the data input and evaluation, without changing the algorithm itself. The specific comparison methods include Group-Sampling, Cluster-Contrast and ICE, and the three methods all use ResNet50-IBN as the backbone network for fair comparison. ICE method does not consider the constraints of cameras. The final test results are shown in Table \ref{tab2}.


\begin{table*}[!t]
\centering
\caption{Basic results}
\label{tab2}
\begin{threeparttable}
\begin{tabular}{|c|cccc|cccc|cccc|}
\hline
\multirow{2}{*}{Methods} & \multicolumn{4}{c|}{CUB-200-2011}  & \multicolumn{4}{c|}{Oxford-Flowers} & \multicolumn{4}{c|}{Oxford-Pets}  \\
                         & Top-1  & ACC    & NMI    & ARI    & Top-1   & ACC    & NMI    & ARI    & Top-1  & ACC    & NMI    & ARI   \\ \hline
\tnote{$\dagger$} Group-Sampling~\cite{DBLP:journals/corr/abs-2107-03024}           & 24.1\% & 13.0\% & 35.3\% & 0.8\%  & 66.4\%  & 29.2\% & 59.2\% & 16.6\% & 18.3\% & 10.8\% & 14.5\% & 0.4\% \\
\tnote{$\dagger$} Cluster-Contrast~\cite{DBLP:journals/corr/abs-2103-11568}         & 47.0\% & 30.2\% & 57.3\% & 3.5\%  & \underline{79.1\%}  & \underline{34.4\%} & \underline{66.7\%} & \underline{22.7\%} & \underline{71.3\%} & \underline{25.3\%} & \underline{40.2\%} & \underline{2.2\%} \\
\tnote{$\dagger$} ICE~\cite{DBLP:journals/corr/abs-2103-16364}                      & \underline{63.9\%} & \underline{47.2\%} & \underline{73.8\%} & \underline{4.5\%}  & 64.9\%  & 9.8\%  & 30.9\% & 0.2\%  & 19.2\% & 4.5\%  & 4.6\%  & 0.0\% \\
Our UFCL                 & \textbf{69.0\%} & \textbf{58.0\%} & \textbf{78.6\%} & \textbf{45.3\%} & \textbf{90.1\%}  & \textbf{39.2\%} & \textbf{70.7\%} & \textbf{26.5\%} & \textbf{79.0\%} & \textbf{28.1\%} & \textbf{44.8\%} & \textbf{3.6\%} \\
\hline
\multirow{2}{*}{Methods} & \multicolumn{4}{c|}{Stanford-Dogs}  & \multicolumn{4}{c|}{Stanford-Cars} & \multicolumn{4}{c|}{FGVC-Aircraft}  \\
                         & Top-1  & ACC    & NMI    & ARI    & Top-1   & ACC    & NMI    & ARI    & Top-1  & ACC    & NMI    & ARI   \\ \hline
\tnote{$\dagger$} Group-Sampling~\cite{DBLP:journals/corr/abs-2107-03024}           & 28.5\%  & 10.0\% & 22.3\% & 0.1\% & 16.8\%  & 9.1\%  & 28.3\% & 0.3\% & 27.3\%  & 6.7\%  & 19.8\% & 0.1\% \\
\tnote{$\dagger$} Cluster-Contrast~\cite{DBLP:journals/corr/abs-2103-11568}         & 40.2\%  & 15.9\% & 31.5\% & 0.3\% & \textbf{22.4\%}  & \textbf{12.0\%} & \textbf{36.3\%} & \textbf{0.4\%} & \textbf{41.2\%}  & \textbf{6.2\%} & \textbf{20.1\%} & \textbf{0.1\%} \\
\tnote{$\dagger$} ICE~\cite{DBLP:journals/corr/abs-2103-16364}                      & \textbf{63.7\%}  & \underline{30.1\%} & \underline{44.9\%} & \underline{0.8\%} & 20.7\%  & 10.0\% & 32.0\% & 0.1\% & 19.6\%  & 2.3\%  & 4.9\%  & 0.0\% \\
Our UFCL                 & \underline{58.5\%}  & \textbf{31.1\%} & \textbf{45.8\%} & \textbf{0.8\%} & \textbf{33.7\%}  & \textbf{19.4\%} & \textbf{46.5\%} & \textbf{3.3\%} & \textbf{43.3\%}  & \underline{6.0\%}  & \underline{19.0\%} & \underline{0.1\%} \\
\hline
\end{tabular}
 \begin{tablenotes}
        \footnotesize
        \item[$\dagger$] The reproduced method for FGVC.
 \end{tablenotes}
\end{threeparttable}
\end{table*}

It can be easily found that our UFCL is the best on the whole in terms of classification Top-1 and clustering ACC. Our UFCL have the best Top-1 except Stanford-Dogs and best ACC except FGVC-Aircraft. Cluster-Contrast is the second best method, which achieves the second best Top-1 performance on four datasets. ICE has unstable performance. It has good results on CUB-200-2011, Stanford-Dogs and Oxford-Flowers, but the results are poor on the other three datasets. Group-Sampling is  poor generally.

Comparing different datasets, it can be found that Oxford-Flowers and Oxford-Pets are relatively easy for classification. Most methods have high classification Top-1 and clustering ACC on these two datasets, where our UFCL achieves 90.1\% and 79.0\% in Top-1. It is relatively difficult for classification on Stanford-Cars and FGVC-Aircraft, in which our UFCL only achieves 33.7\% and 43.3\% in Top-1. In terms of clustering ACC, our UFCL has the best result on CUB-200-2011, reaching 58.0\% in ACC, followed by Oxford-Flowers and Stanford-Dogs, reaching 39.2\% and 31.1\% in ACC respectively. It has the worst performance on FGVC-Aircraft, on which different methods are similar ACC. It can be seen that it is difficult to do unsupervised learning on FGVC-Aircraft.

Based on the aforementioned results, it can be found that the unsupervised learning methods of person re-ID has great gaps when they are applied to fine-grained recognition tasks, meanwhile there are also great performance difference on different datasets.

\subsection{Ablation Analysis}
\label{ablation}

\subsubsection{Feature Extraction Module}

In order to measure the feature extraction capability of different networks, we use different feature extraction networks, ResNet-18, ResNet50, MobileNetV2, DenseNet121, ResNet50-IBN respectively. The experiments are conducted on CUB-200-2011 and Oxford-Flowers, and the results are shown in Table \ref{tab3}. Due to the limitation of the CUDA memory, the corresponding batch-size is adjusted from 256 to 192 when DenseNet121 is used.

\begin{table*}[!t]
\centering
\caption{Results of different feature extraction networks.}
\label{tab3}
\begin{tabular}{|c|c|cccc|cccc|}
\hline
\multirow{2}{*}{Backbone} & \multirow{2}{*}{Batch Size} & \multicolumn{4}{c|}{CUB-200-2011}  & \multicolumn{4}{c|}{Oxford-Flowers} \\
                          &                             & Top-1  & ACC    & NMI    & ARI    & Top-1   & ACC    & NMI    & ARI    \\
                          \hline
MobileNetV2               & 256                         & 55.6\% & 46.1\% & 71.4\% & 32.2\% & 88.4\%  & 38.5\% & 70.2\% & 26.3\% \\
DenseNet121               & 192                         & \underline{62.3\%} & \underline{48.6\%} & \underline{72.8\%} & \underline{28.3\%} & \textbf{90.8\%}  & \textbf{39.5\%} & \textbf{71.3\%} & \textbf{27.3\%} \\
ResNet18                  & 256                         & 39.3\% & 34.8\% & 64.1\% & 24.4\% & 81.5\%  & 36.3\% & 67.4\% & 23.6\% \\
ResNet50                  & 256                         & 44.9\% & 37.9\% & 65.8\% & 25.5\% & 88.0\%  & 37.6\% & 69.7\% & 25.2\% \\
ResNet50-IBN              & 256                         & \textbf{69.0\%} & \textbf{58.0\%} & \textbf{78.6\%} & \textbf{45.3\%} & \underline{90.1\%}  & \underline{39.2\%} & \underline{70.7\%} & \underline{26.5\%} \\
\hline
\end{tabular}
\end{table*}

It can be found from Table \ref{tab3} that ResNet50-IBN has the best performance on CUB-200-2011, reaching 69.0\% in Top-1 and 58.0\% in ACC. Secondly, DenseNet121 reached 62.3\% in Top-1 and 48.6\% in ACC. The Top-1 and ACC of ResNet-18 were 39.3\% and 34.8\% respectively, which have 29.7\% and 23.3\% gaps from the best ResNet50-IBN. It can also be found that the light-weighted MobileNetV2 has achieved better results than ResNet50, reaching 55.6\% in Top-1 and 46.1\% in ACC. The main reason is that MobileNetV2 is proposed by absorbing the advantage of ResNet architecture and has better generalization ability.

\begin{figure}[!t]
\centering
\subfigure[]{
\label{fig5a}
\includegraphics[width=0.23\textwidth]{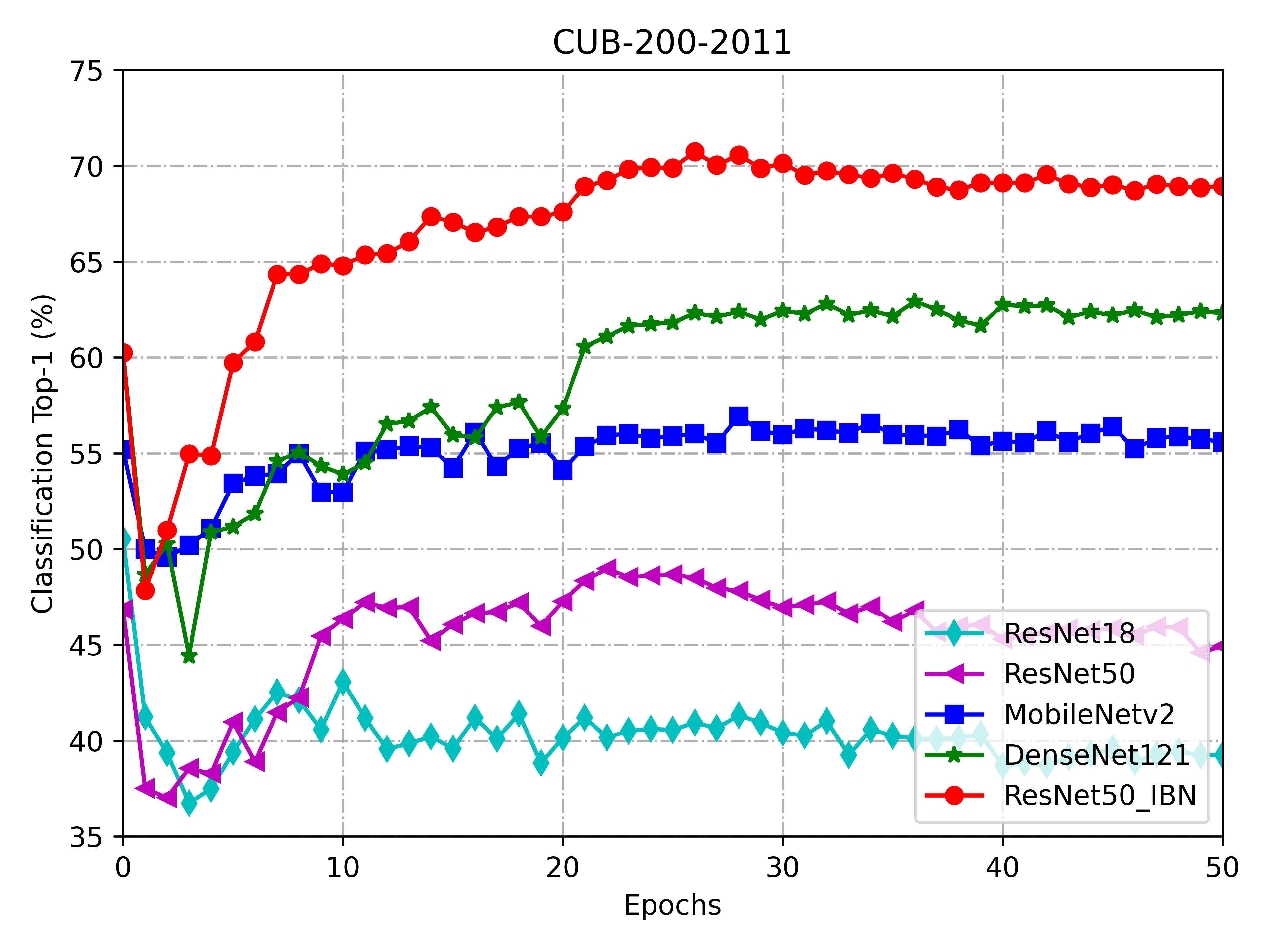}}
\subfigure[]{
\label{fig5b}
\includegraphics[width=0.23\textwidth]{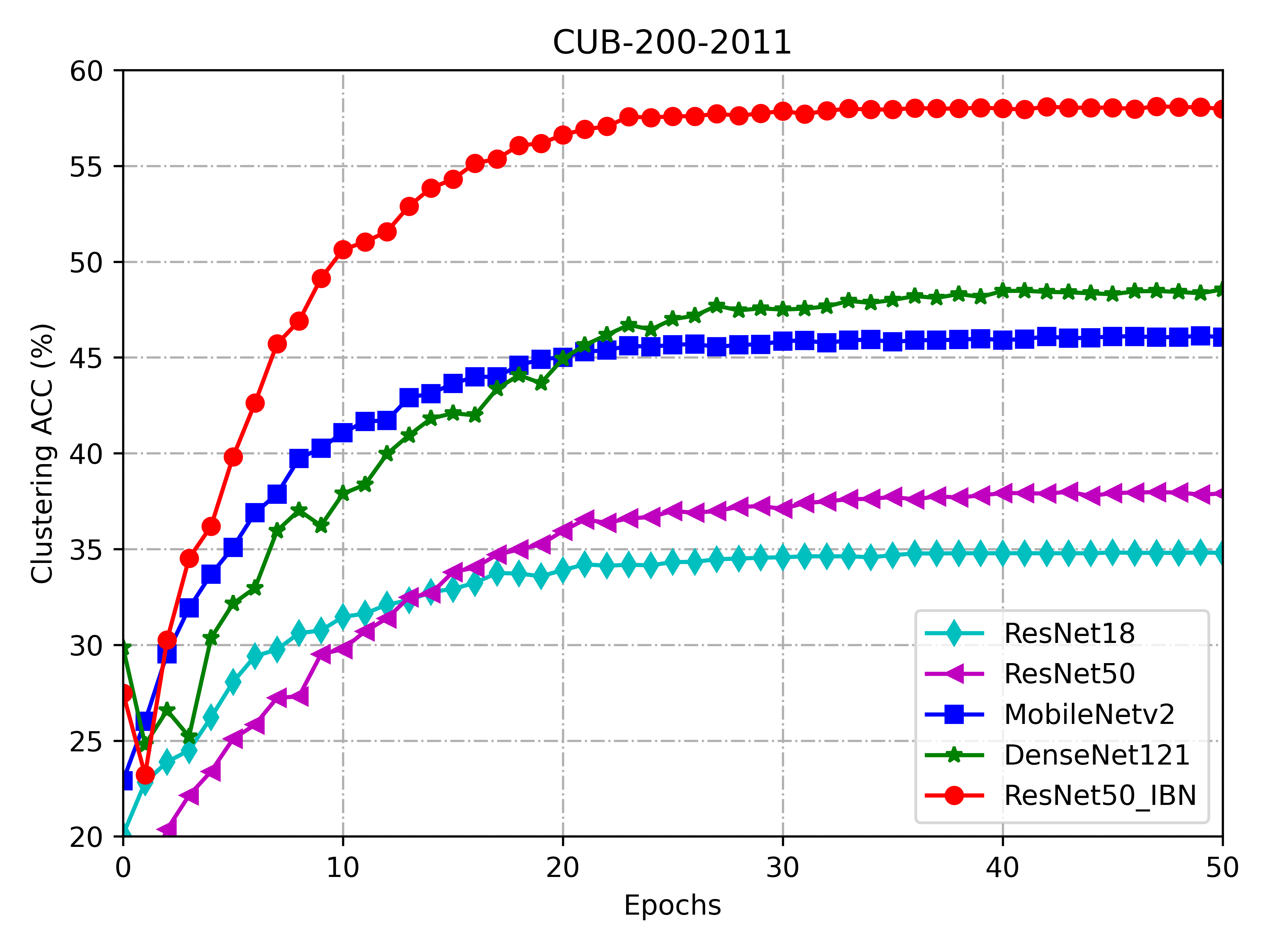}}
\subfigure[]{
\label{fig5c}
\includegraphics[width=0.23\textwidth]{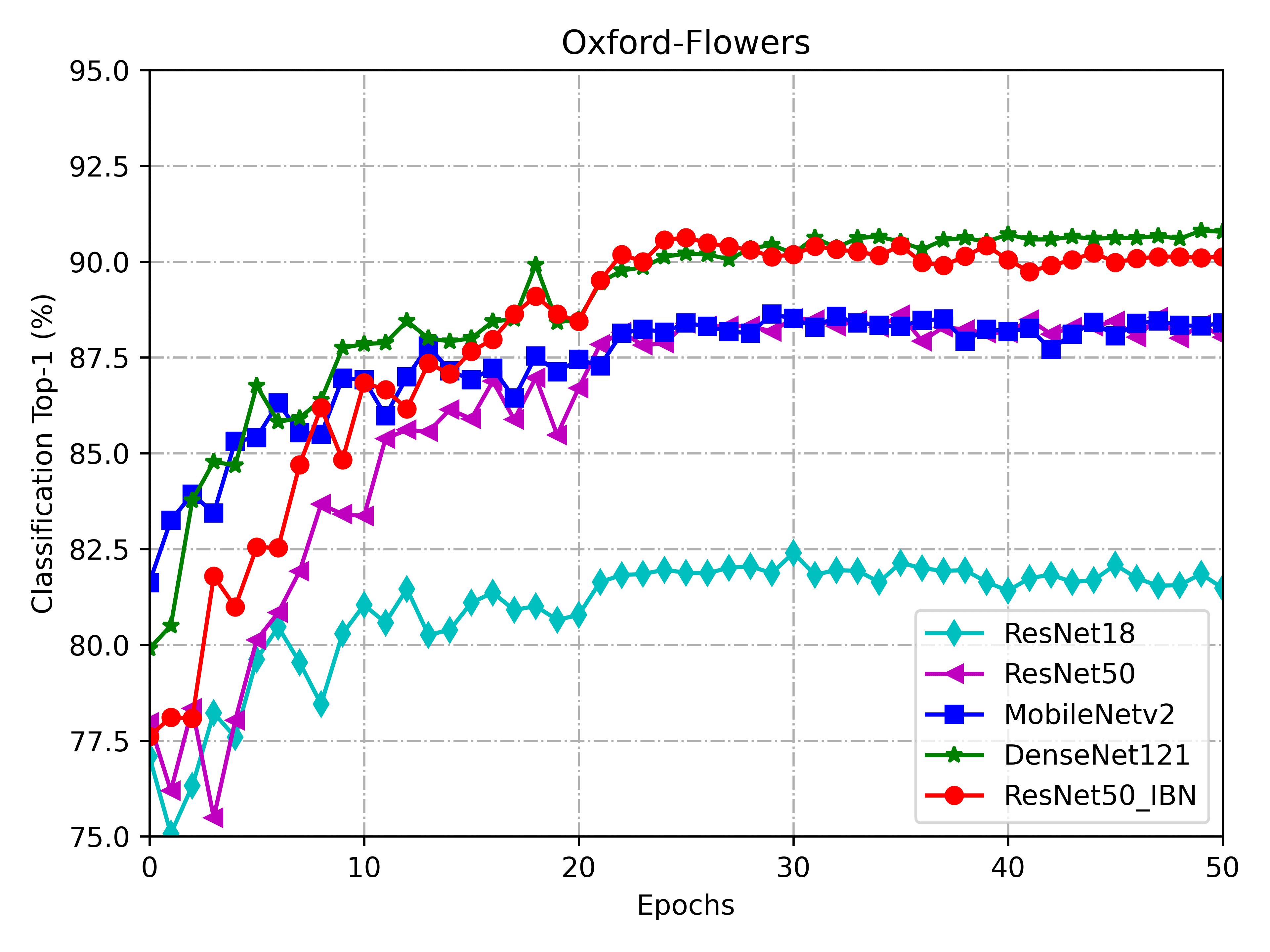}}
\subfigure[]{
\label{fig5d}
\includegraphics[width=0.23\textwidth]{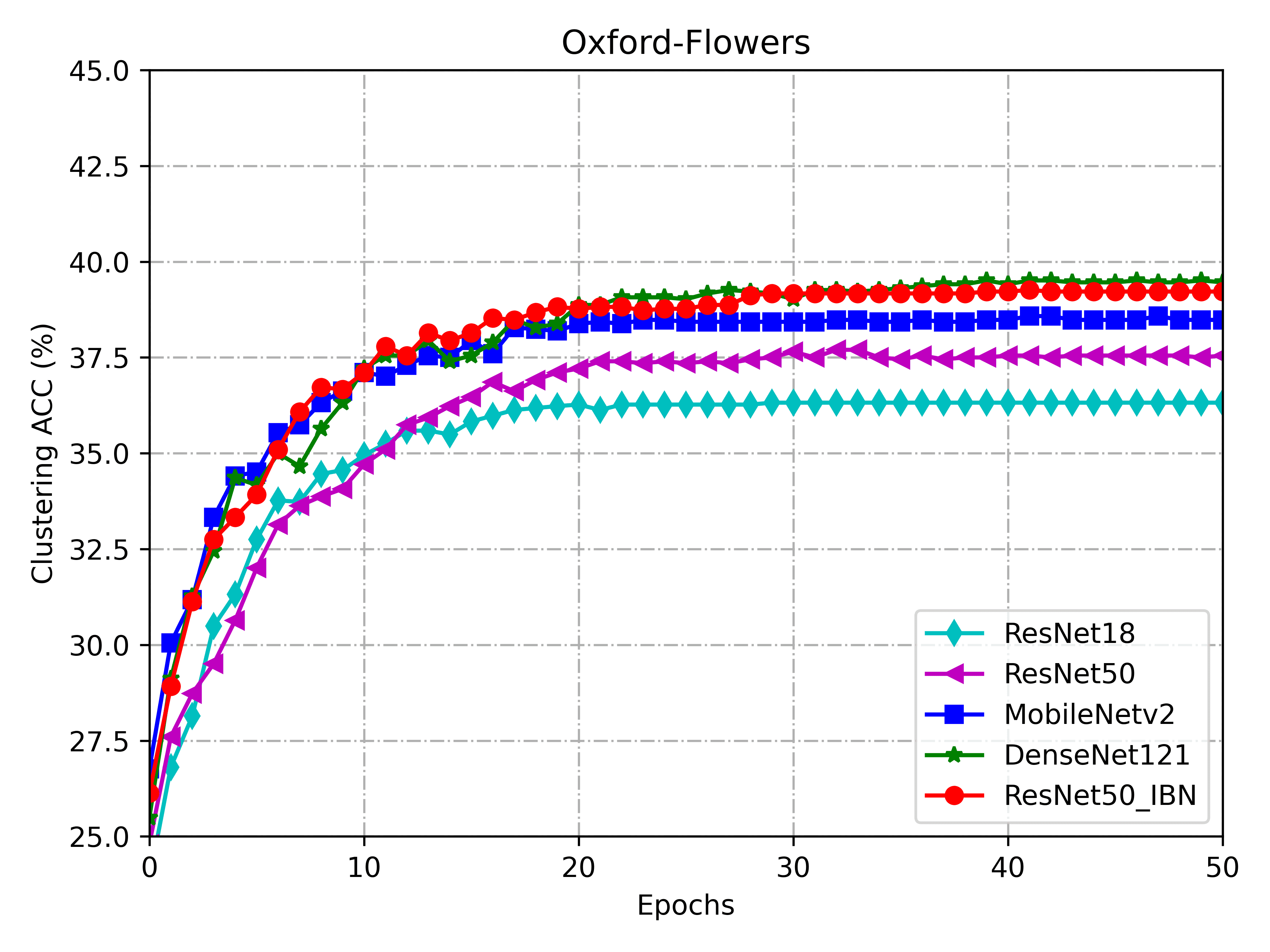}}
\caption{Performance curves of different feature extraction networks.}
\label{fig5}
\end{figure}

At the same time, the parameters of feature extraction network are initialized by the pre-trained model on the ImageNet dataset, so the initial feature distribution is affected by the prior knowledge of ImageNet dataset and has an important impact on the results of generating pseudo labels by the first clustering. The results of different networks is evaluated on CUB-200-2011 and Stanford-Flowers and shown in Fig.~\ref{fig5}. Fig.~\ref{fig5} shows the curves of Top-1 and ACC in training process. It can be found that most of networks have relatively good initial classification Top-1, but it has a relatively large decline after the first epoch. This is mainly because the initialized parameters need an adaptation process on a new dataset. For the clustered ACC, the initial feature distribution is relatively scattered, and the value is relatively low. After several epochs, the performance is also continuously improved. In addition, comparing different feature extraction networks, it can be found that a small-scale network (ResNet18) is not as good as a large-scale network (ResNet50).

\subsubsection{Hierarchical Clustering Module}
\begin{figure}[!t]
\centering
\subfigure[]{
\label{fig6a}
\includegraphics[width=0.23\textwidth]{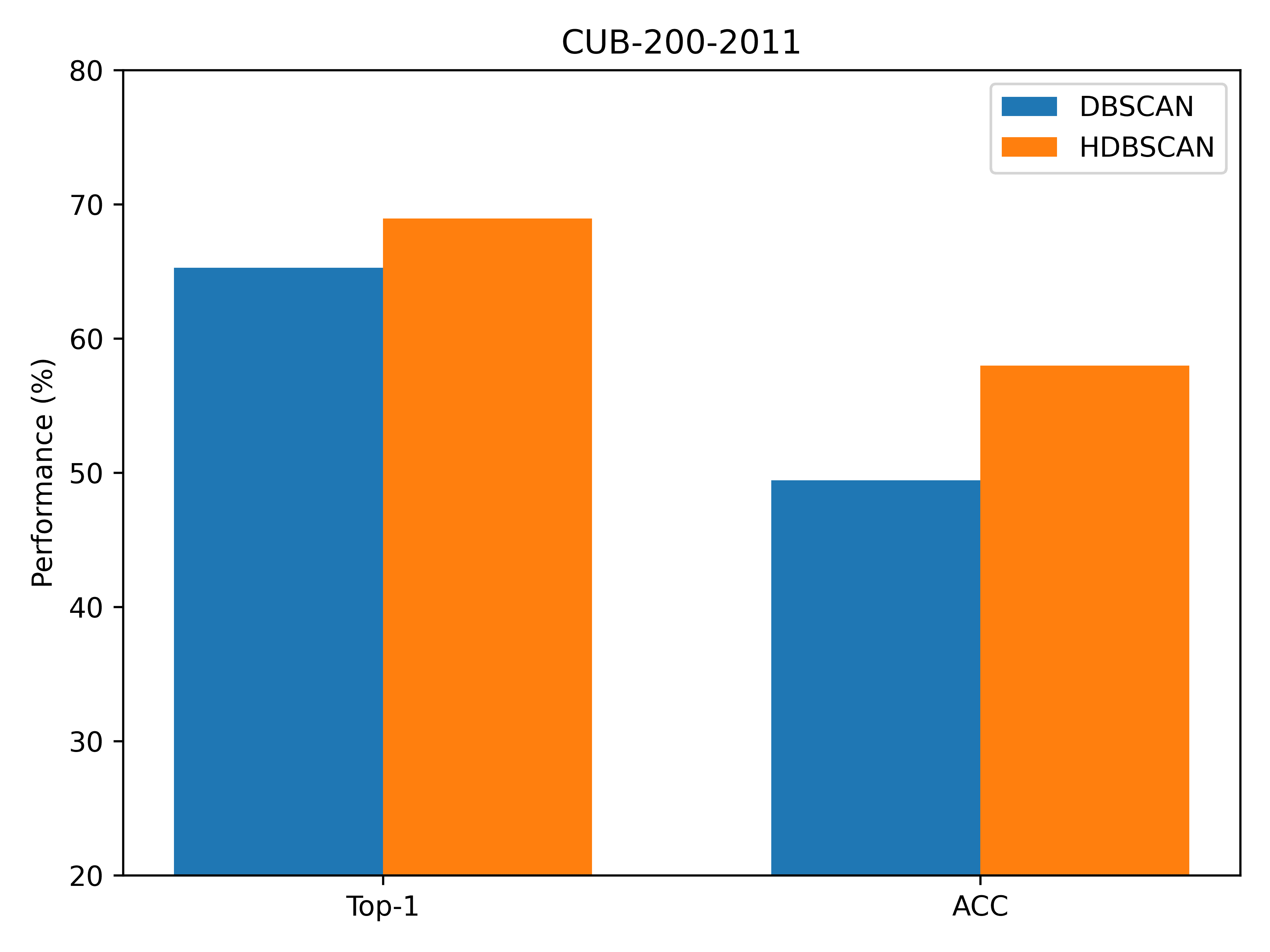}}
\subfigure[]{
\label{fig6b}
\includegraphics[width=0.23\textwidth]{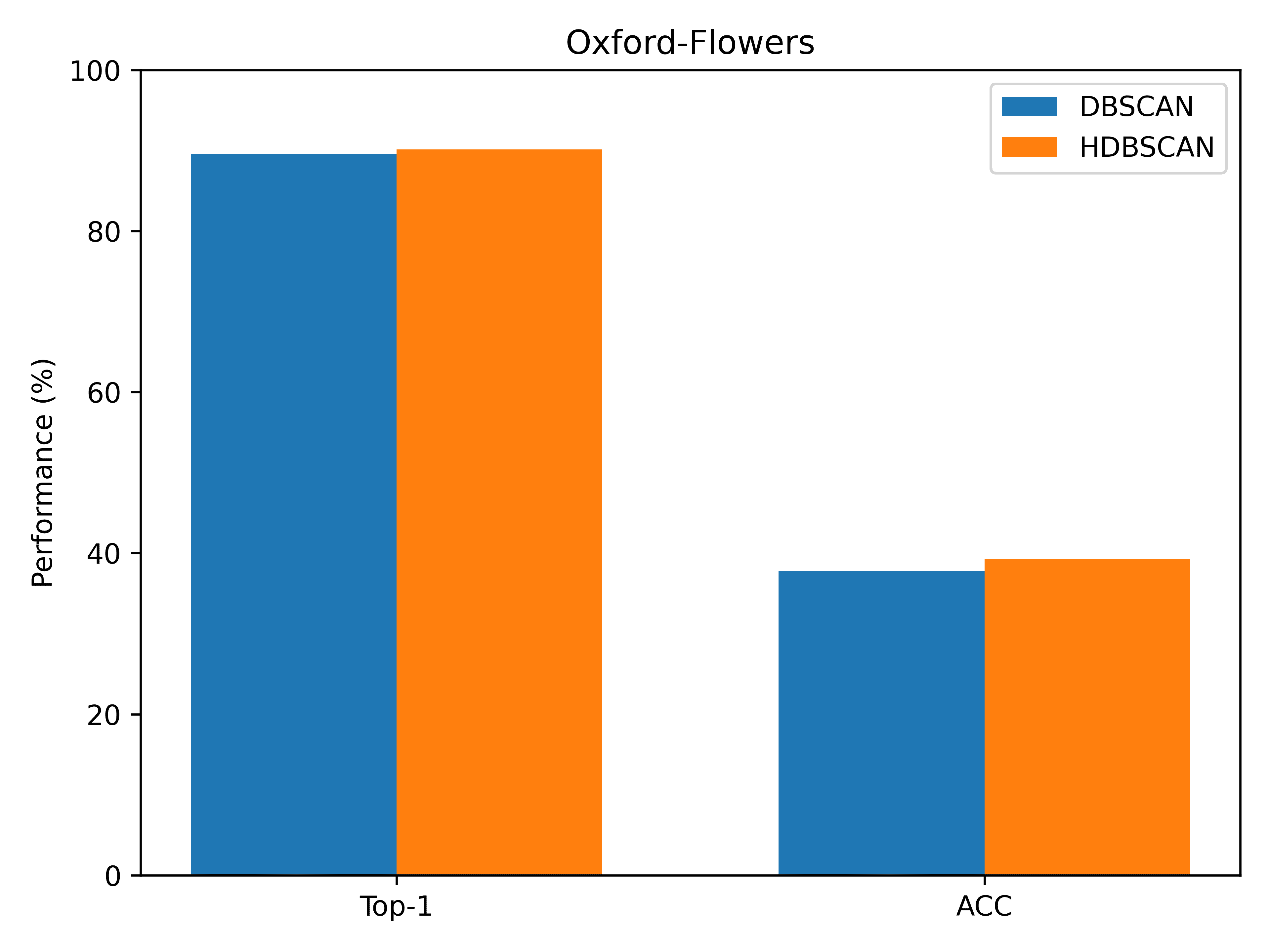}}
\caption{Comparison of DBSCAN and HDBSCAN.}
\label{fig6}
\end{figure}

In order to verify the effect of HDBSCAN, the widely used DBSCAN algorithm is compared, in which the DBSCAN parameter $\varepsilon $ is uniformly set to 0.4 and the number of adjacent examples is set to 4. The experimental results are shown in Fig.~\ref{fig6}.

Compared with DBSCAN, HDBSCAN has increased by 3.7\% and 0.5\% in Top-1, and 8.6\% and 1.4\% in ACC on CUB-200-2011 and Oxford-Flowers, respectively. It is verified that HDBSCAN produces better clustering results for unsupervised learning. From another viewpoint, the ACC of HDBSCAN has a big gap between CUB-200-2011 and Oxford-Flowers. It indicates that HDBSCAN has better clustering results on CUB-200-2011 than on Oxford-Flowers. The Top-1 of HDBSCAN has also a big gap between CUB-200-2011 and Oxford-Flowers. However, it has higher Top-1 on Oxford-Flowers than that on CUB-200-2011. It indicates that HDBSCAN has better classification results on Oxford-Flowers than that on CUB-200-2011. Comparing ACC and Top-1, it can be found that the clustering effectiveness are not necessarily positively correlated with the classification accuracy. In addition, HDBSCAN has only one parameter, which has better applicability.

\subsubsection{Contrastive Learning Module}
The feature agent proposed in this paper is similar to Cluster-Contrast, which is the feature center of the examples in the same cluster. The difference is that our method is a generalization of that in Cluster-Contrast. When we use \emph{zero distance}, it is the same as the method in Cluster-Contrast. At the same time, our method directly uses examples to update feature agents, unlike Cluster-Contrast, which uses difficult examples in batch examples to update feature agents. In addition, in order to verify the advantages of the updating strategy adopted in this paper, we also compare it with Group-Sampling, which maintains the feature of each example in memory to do updating. In an execution, it updates the feature of a single example. The experimental results are shown in Table \ref{tab4}.

\begin{table*}[!t]
\centering
\caption{Results of different updating strategies.}
\label{tab4}
\begin{tabular}{|c|cccc|cccc|}
\hline
\multirow{2}{*}{Methods} & \multicolumn{4}{c|}{CUB-200-2011}  & \multicolumn{4}{c|}{Oxford-Flowers} \\
                         & Top-1  & ACC    & NMI    & ARI    & Top-1   & ACC    & NMI    & ARI    \\ \hline
Group-Sampling~\cite{DBLP:journals/corr/abs-2107-03024}           & 24.1\% & 13.0\% & 35.3\% & 0.8\%  & 66.4\%  & 29.2\% & 59.2\% & 16.6\% \\
Cluster-Contrast~\cite{DBLP:journals/corr/abs-2103-11568}         & 47.0\% & 30.2\% & 57.3\% & 3.5\%  & 79.1\%  & 34.4\% & 66.7\% & 22.7\% \\
Our UFCL (zero)           & 67.1\% & 56.7\% & 77.8\% & 43.9\% & \underline{87.9\%} & 37.5\% & 69.7\% & 25.5\% \\
Our UFCL (min)            & \underline{68.2\%} & \underline{57.0\%} & \underline{78.6\%} & \underline{44.7\%} & 87.5\%  & \underline{38.1\%} & \underline{70.2\%} & \underline{26.2\%} \\
Our UFCL (mean)           & \textbf{69.0\%} & \textbf{58.0\%} & \textbf{78.6\%} & \textbf{45.3\%} & \textbf{90.1\%}  & \textbf{39.2\%} & \textbf{70.7\%} & \textbf{26.5\%} \\ \hline
\end{tabular}
\end{table*}

It can be found from Table \ref{tab4} that on CUB-200-2011 and Oxford flowers, the three weighted strategies proposed in this paper are significantly improved on Top-1 and ACC compared with Cluster-Contrast and Group-Sampling. This improvement is mainly caused by the proposed distance calculation strategy. At the same time, there are some difference among the three different distance measures. The \emph{mean distance} is the best one.

\subsection{Factor Analysis}

In this paper, the pooling strategy in feature extraction, the hyper-parameter in HDBSCAN and the iterations of a single epoch have a great impact on the results. The specific analysis is as follows:

\subsubsection{GEM Pooling}
To test the effectiveness of GEM, we compare it with the widely used GAP and GMP. On CUB-200-2011 and Oxford-Flowers, all experimental conditions and parameter settings were the same, and only the pooling method was different. Four methods including GAP, GMP, GAP+GMP and GEM were compared. The results are shown in Table \ref{tab5}.

\begin{table*}[!t]
\centering
\caption{Results of different pooling operations.}
\label{tab5}
\begin{tabular}{|c|cccc|cccc|}
\hline
\multirow{2}{*}{Pooling} & \multicolumn{4}{c|}{CUB-200-2011}  & \multicolumn{4}{c|}{Oxford-Flowers} \\
                         & Top-1  & ACC    & NMI    & ARI    & Top-1   & ACC    & NMI    & ARI    \\ \hline
GAP                      & 44.1\% & 35.6\% & 63.8\% & 24.1\% & 89.2\%  & 37.7\% & 70.1\% & 25.8\% \\
GMP                      & 68.2\% & 60.1\% & 79.9\% & 47.8\% & 87.5\%  & 38.2\% & 69.4\% & 25.3\% \\
GAP+GMP                  & \underline{68.4\%} & \textbf{61.6\%} & \textbf{80.7\%} & \textbf{50.0\%} & \underline{88.6\%}  & \underline{38.2\%} & \underline{69.9\%} & \underline{25.7\%} \\
GEM                      & \textbf{69.0\%} & \underline{58.0\%} & \underline{78.6\%} & \underline{45.3\%} & \textbf{90.1\%}  & \textbf{39.2\%} & \textbf{70.7\%} & \textbf{26.5\%} \\ \hline
\end{tabular}
\end{table*}

It can be found that under ResNet-50-IBN, the Top-1 of GEM with a parameter reaches 60.0\% and 90.1\% respectively on CUB-200-2011 and Oxford-Flowers datasets, and the ACC reaches 58.0\% and 39.2\% respectively. Compared with GAP, GEM increased Top-1 by 22.4\% and 0.9\%, and increased ACC by 22.4\% and 1.5\% on CUB-200-2011 and Oxford-Flowers respectively. Compared with GMP, it also has a certain improvement. At the same time, the effect is slightly better than that of GAP+GMP. Therefore, GEM pooling has better performance and is good at unsupervised fine-grained classification.

\subsubsection{Super-parameter, min-cluster-size}
\begin{figure}[!t]
\centering
\subfigure[]{
\label{fig7a}
\includegraphics[width=0.23\textwidth]{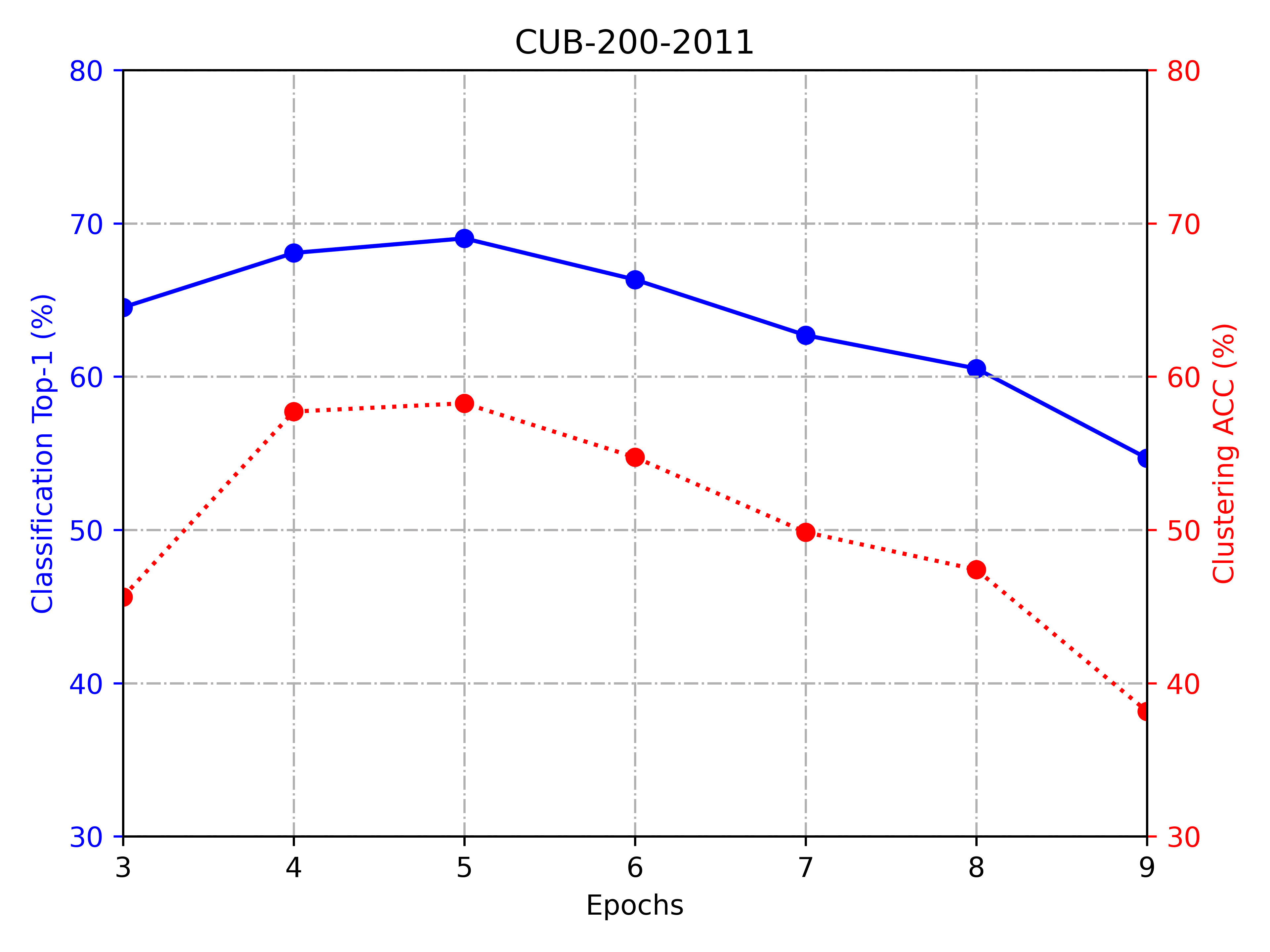}}
\subfigure[]{
\label{fig7b}
\includegraphics[width=0.23\textwidth]{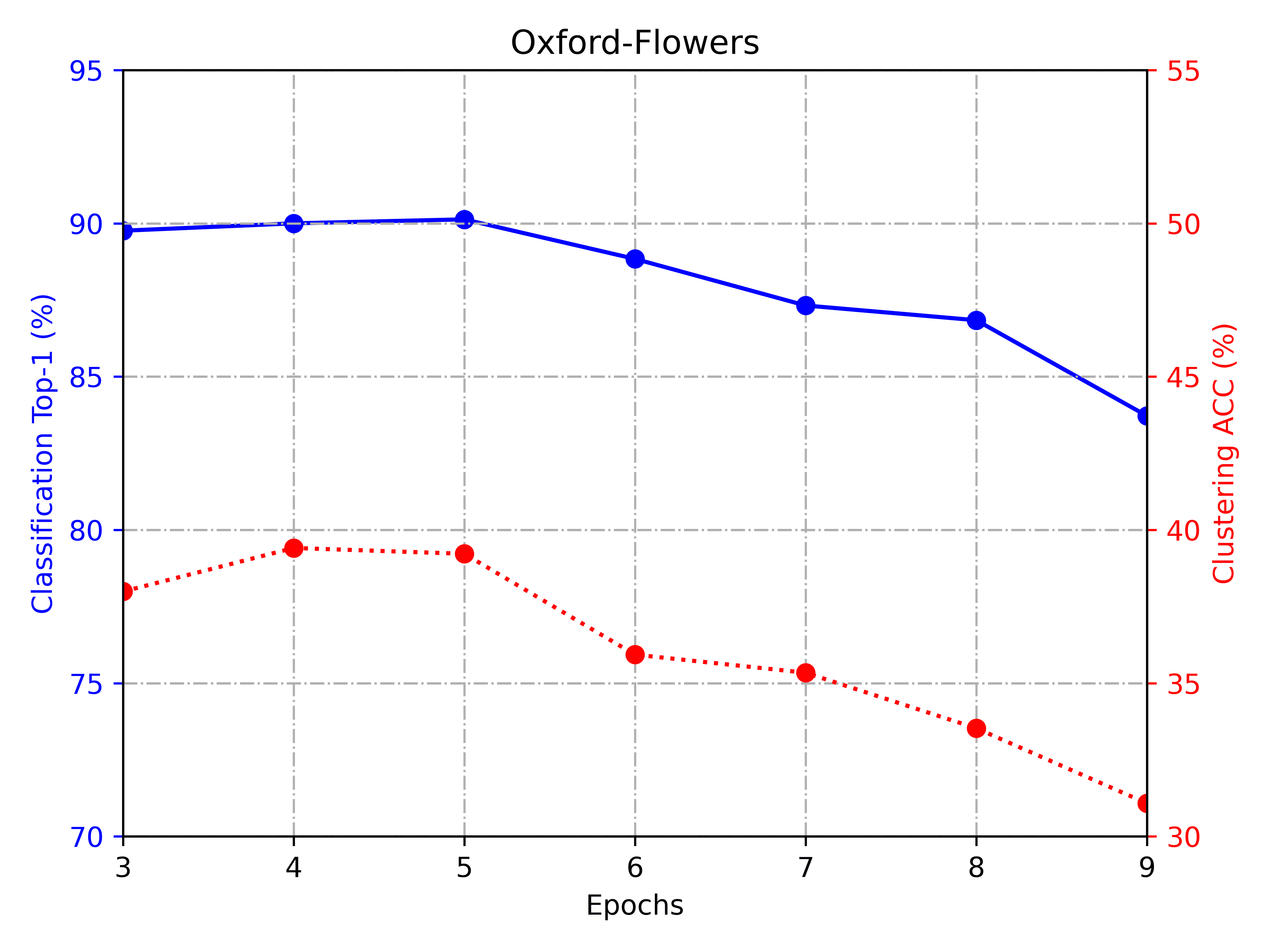}}
\caption{Results of different super-parameters.}
\label{fig7}
\end{figure}

HDBSCAN has a hyper-parameter \emph{min-cluster-size}. We conducted experiments to test the results of 3-11 different values on CUB-200-2011 and Oxford-Flowers. The results are shown in Fig.~\ref{fig7}.

It can be found that the number of clusters will decrease with the increase of parameter value. At the same time, the classification Top-1 and clustering ACC are better when they are close to the actual number of categories. It is difficult to set the optimal clustering parameters without any prior conditions. However, the experimental results show that on CUB-200-2011, when the number of clusters is close to 200, both Top-1 and ACC achieve good results. On Oxford flowers, when the number of clusters is close to 102, Top-1 and ACC are also the best. Therefore, if the number of categories is known in advance, the parameter that makes the number of clusters closest to the number of categories can be selected according to the number of categories.

\subsubsection{Iterations in Each Epoch}

Because each epoch performs many iterations internally to learn a relatively better model for clustering, the number of iterations in a single epoch is a key parameter. Too many iterations will lead to over-fitting, and too few iterations will lead to under-fitting. Therefore, an appropriate number of iterations will have a great impact on performance. We performed 25, 50, 100 and 200 iterations on CUB-200-2011 and Oxford-Flowers respectively. The results are shown in Table \ref{tab6}.

\begin{table*}[!t]
\centering
\caption{Results of different iteration times in one epoch}
\label{tab6}
\begin{tabular}{|c|cccc|cccc|}
\hline
\multirow{2}{*}{Iterations} & \multicolumn{4}{c|}{CUB-200-2011}  & \multicolumn{4}{c|}{Oxford-Flowers} \\
                            & Top-1  & ACC    & NMI    & ARI    & Top-1   & ACC    & NMI    & ARI    \\ \hline
25                          & 68.3\% & 48.7\% & 70.5\% & 7.5\%  & \textbf{90.1\%}  & \textbf{39.2\%} & \textbf{70.7\%} & \textbf{26.5\%} \\
50                          & \textbf{71.0\%} & \underline{58.1\%} & 77.6\% & 35.2\% & \underline{82.1\%}  & \underline{36.8\%} & \underline{68.6\%} & \underline{24.4\%} \\
100                         & \underline{69.0\%} & 58.0\% & \underline{78.6\%} & \underline{45.3\%} & 81.2\%  & 36.2\% & 68.2\% & 23.2\% \\
200                         & 62.6\% & \textbf{58.8\%} & \textbf{79.4\%} & \textbf{46.8\%} & 76.1\%  & 35.2\% & 65.5\% & 17.8\% \\ \hline
\end{tabular}
\end{table*}

It can be found from Table \ref{tab6} that for CUB-200-2011, 50 iterations have the best performance. For Oxford-Flowers, 25 iterations performed best. If iterations are less, the learning in one epoch will be insufficient, and the model is not enough to extract effective features to produce good clusters in the next clustering. If iterations are more, the model will over-fit the training data, the model is difficult to have good generalization, and the clusters generated in the next clustering will be worse.

\subsection{Feature Visualization and Analysis}
\begin{figure*}[!t]
\centering
\subfigure[]{
\label{fig8a}
\includegraphics[width=0.18\textwidth]{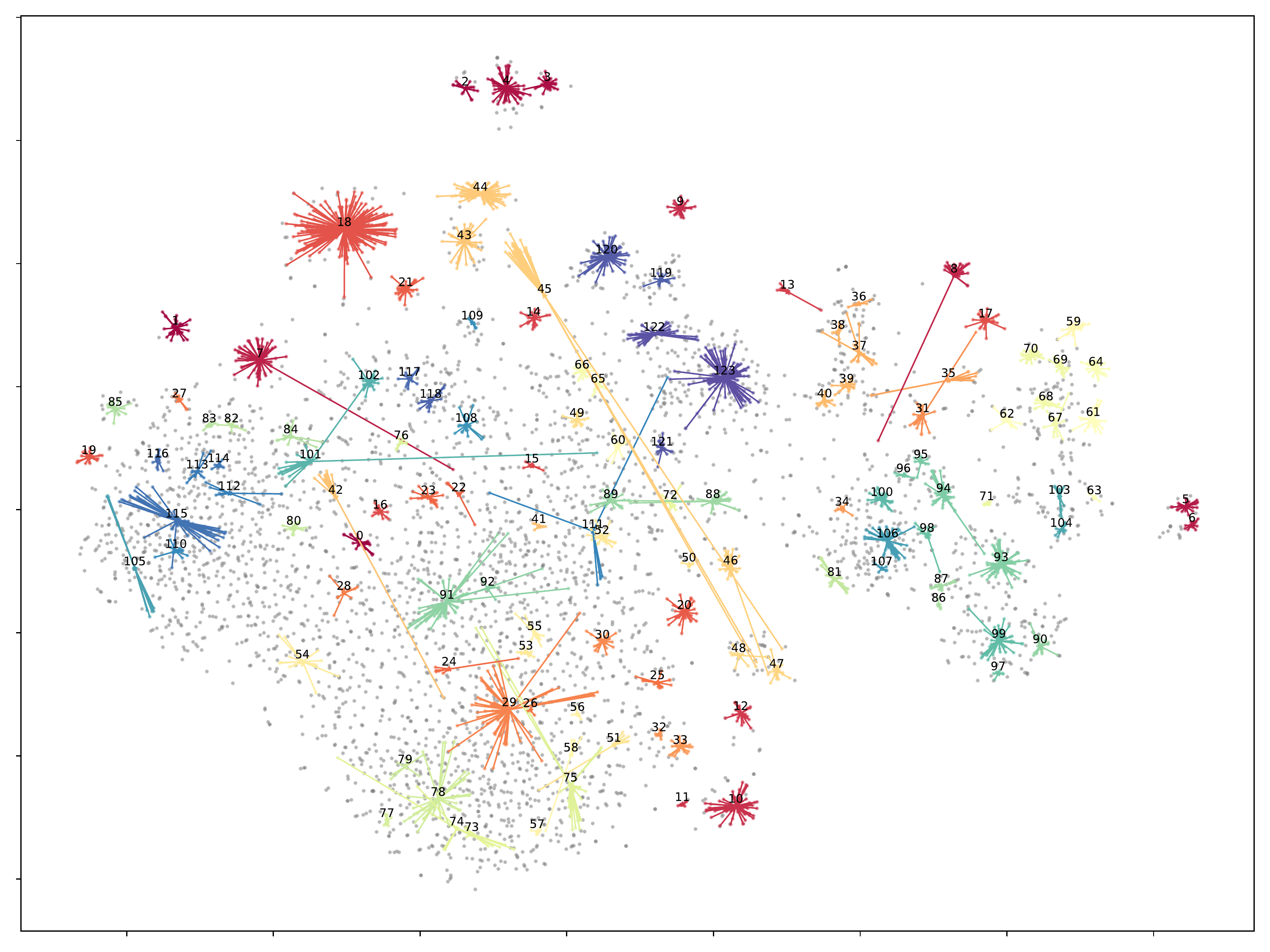}}
\subfigure[]{
\label{fig8b}
\includegraphics[width=0.18\textwidth]{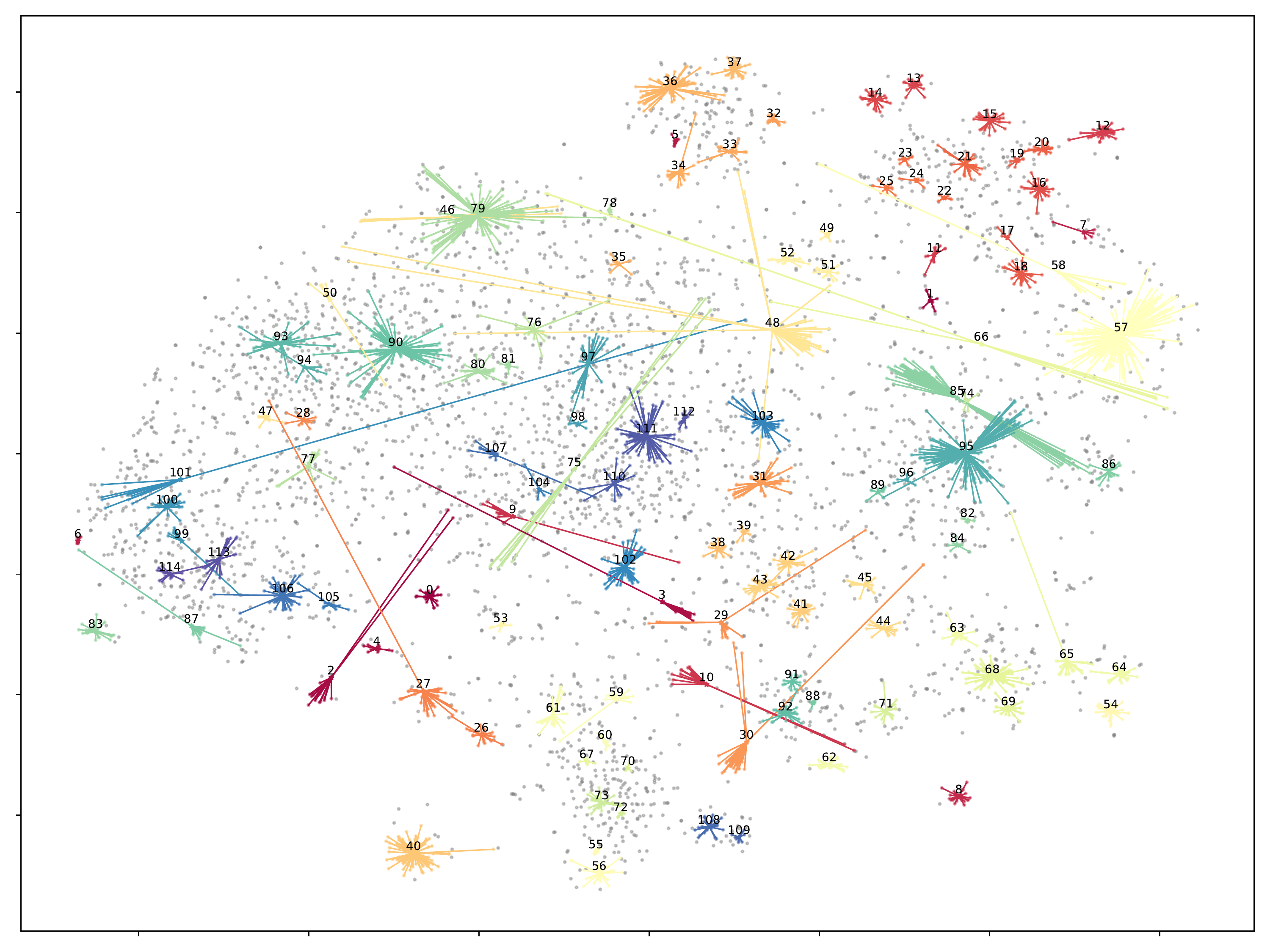}}
\subfigure[]{
\label{fig8c}
\includegraphics[width=0.18\textwidth]{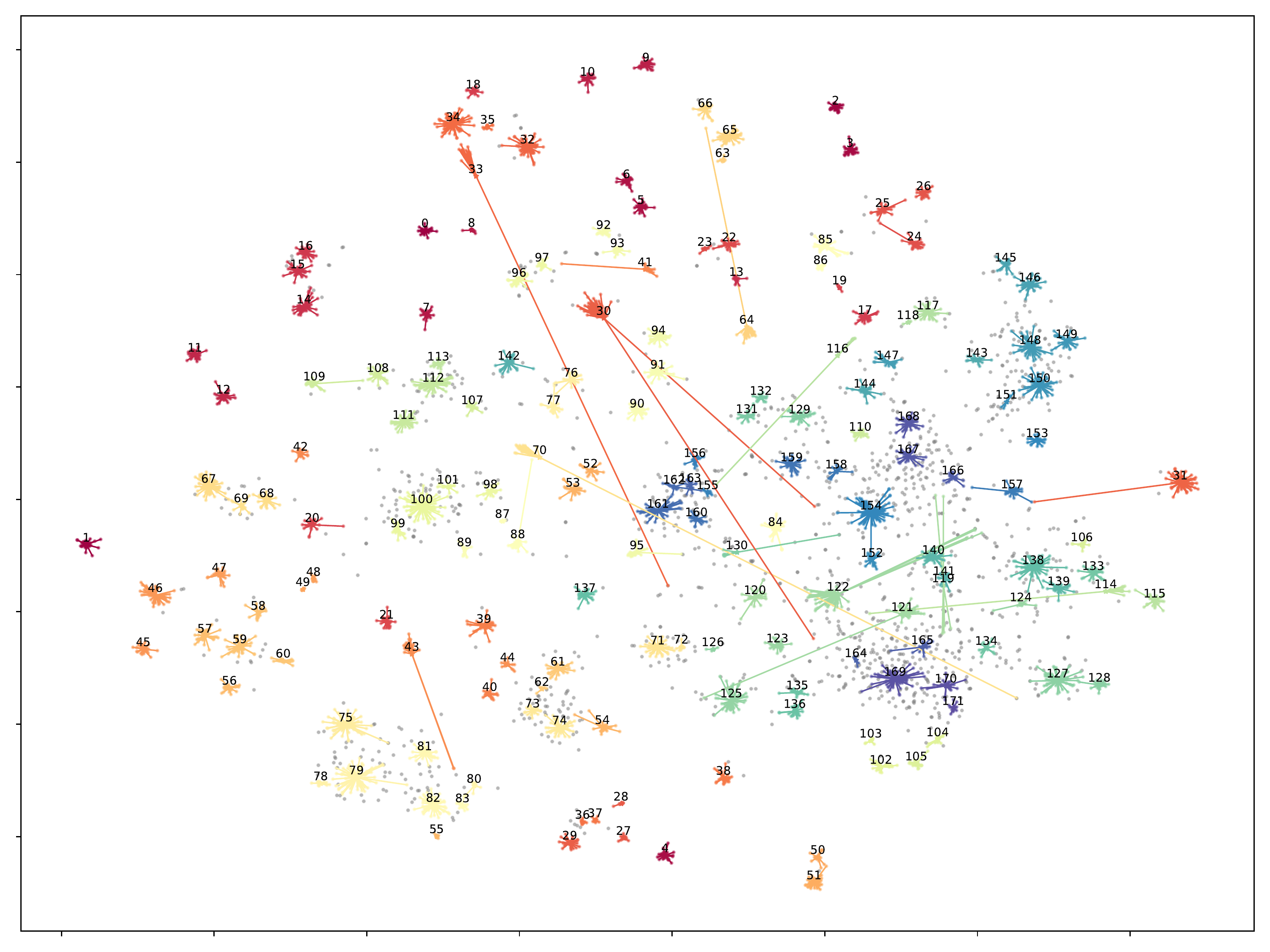}}
\subfigure[]{
\label{fig8d}
\includegraphics[width=0.18\textwidth]{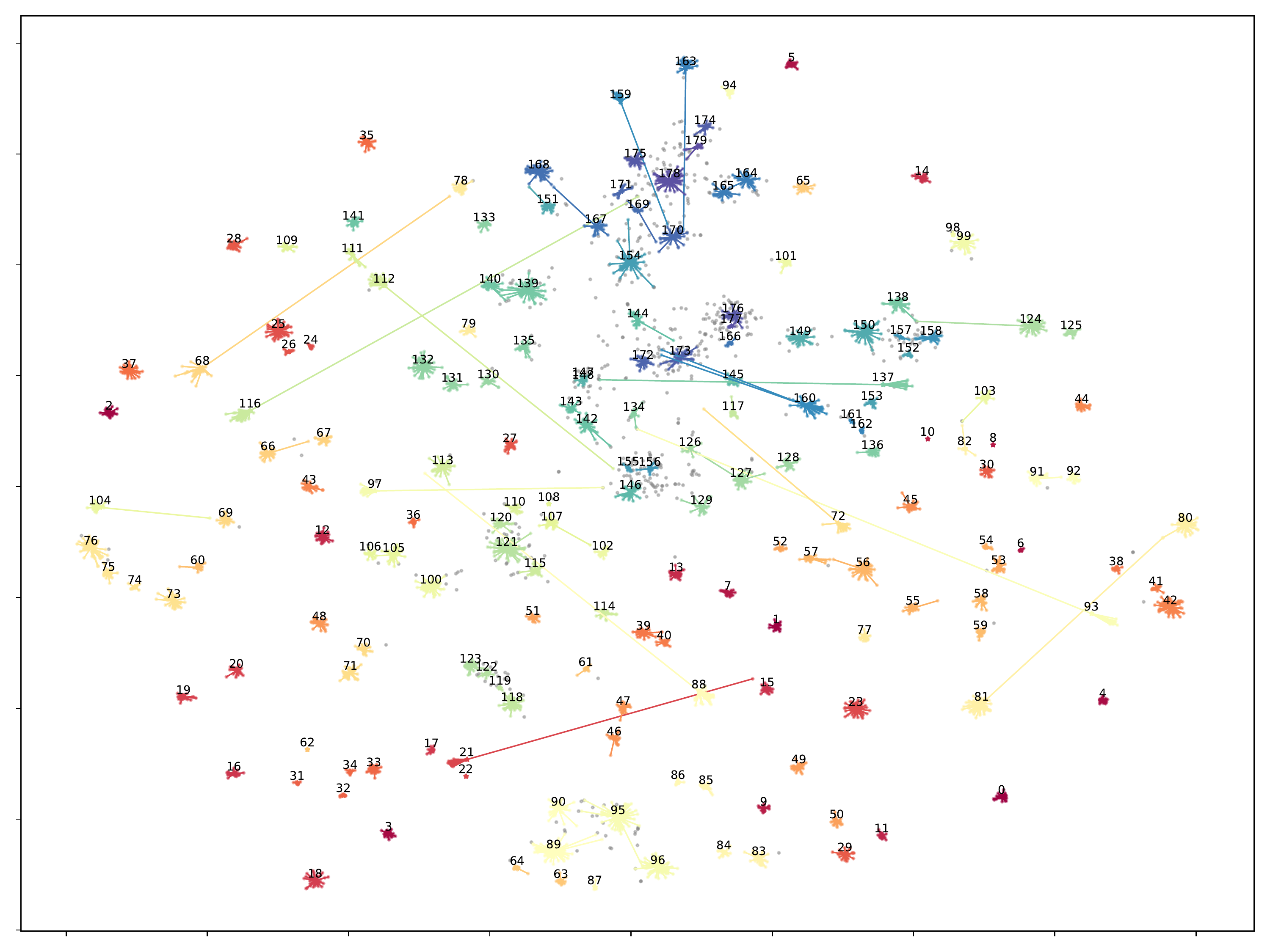}}
\subfigure[]{
\label{fig8e}
\includegraphics[width=0.18\textwidth]{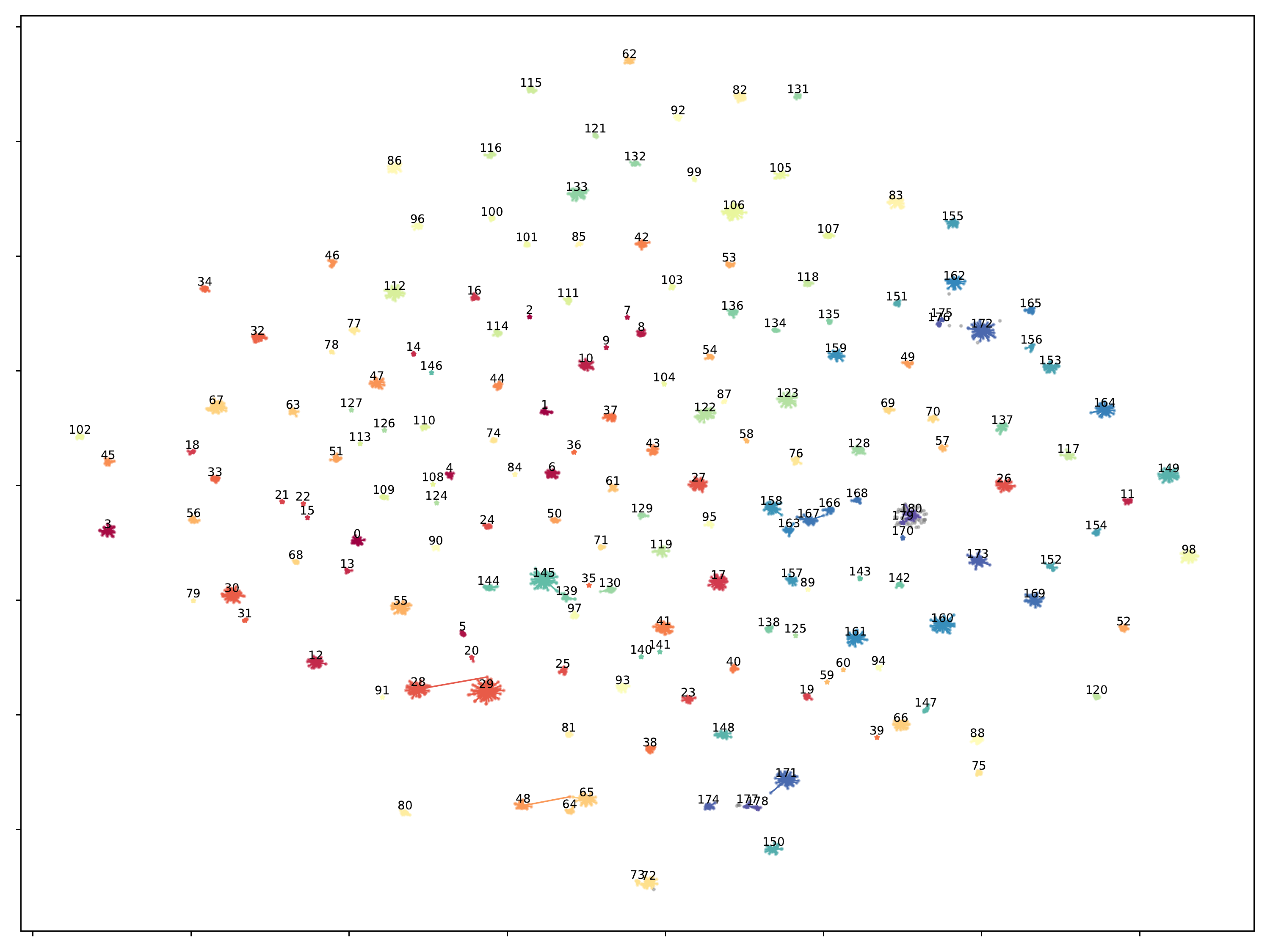}}
\subfigure[]{
\label{fig8a1}
\includegraphics[width=0.18\textwidth]{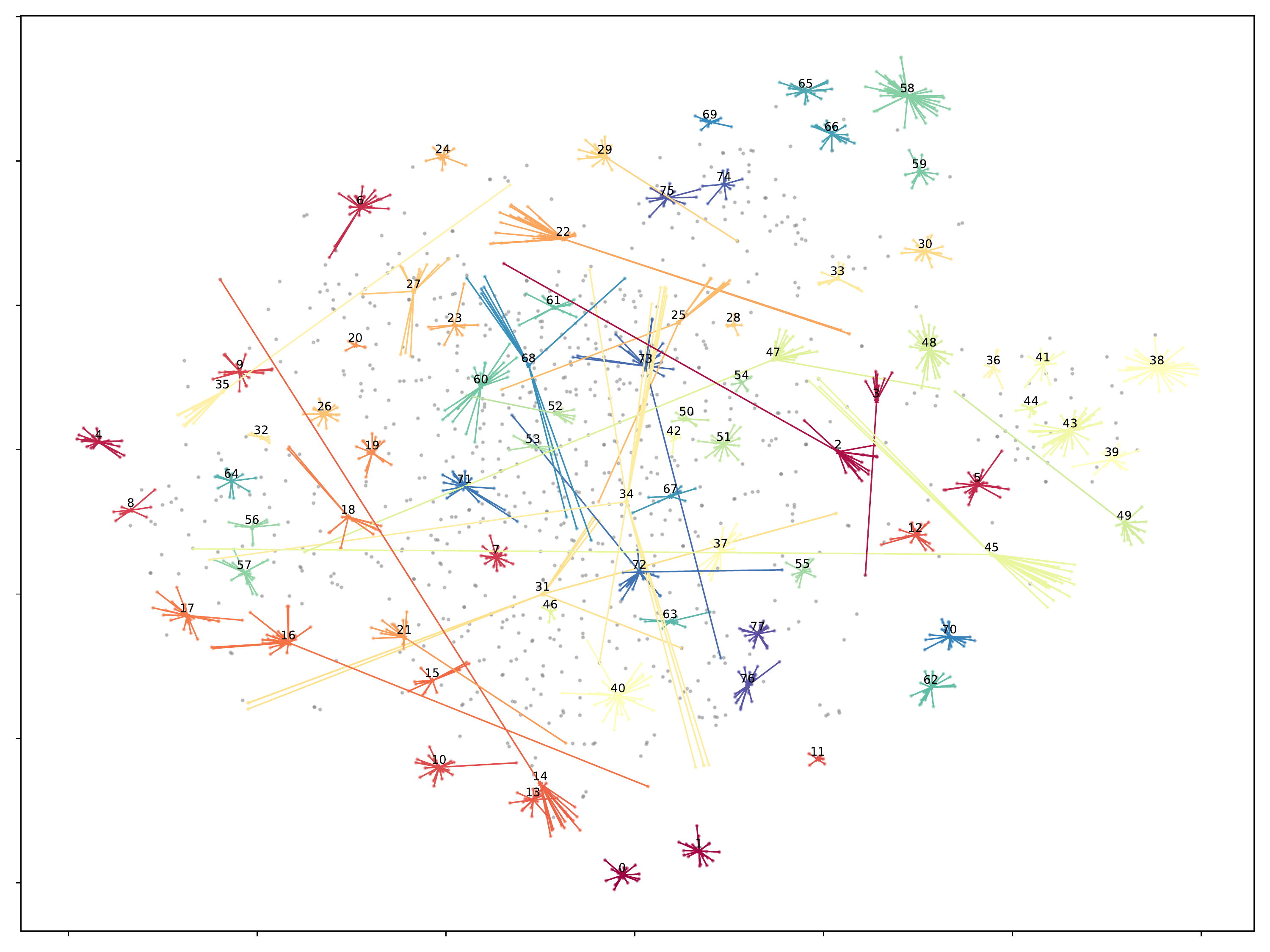}}
\subfigure[]{
\label{fig8b1}
\includegraphics[width=0.18\textwidth]{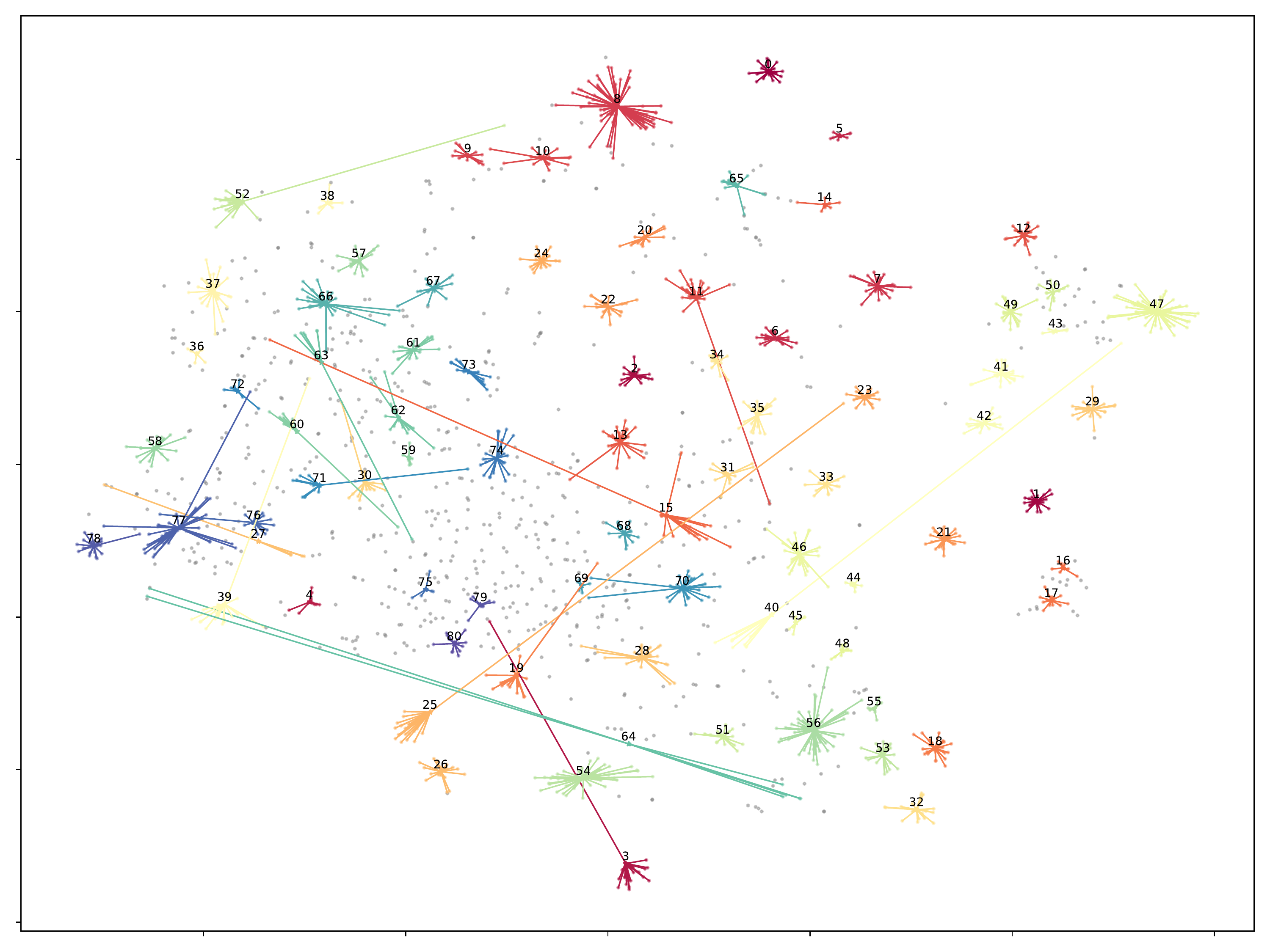}}
\subfigure[]{
\label{fig8c1}
\includegraphics[width=0.18\textwidth]{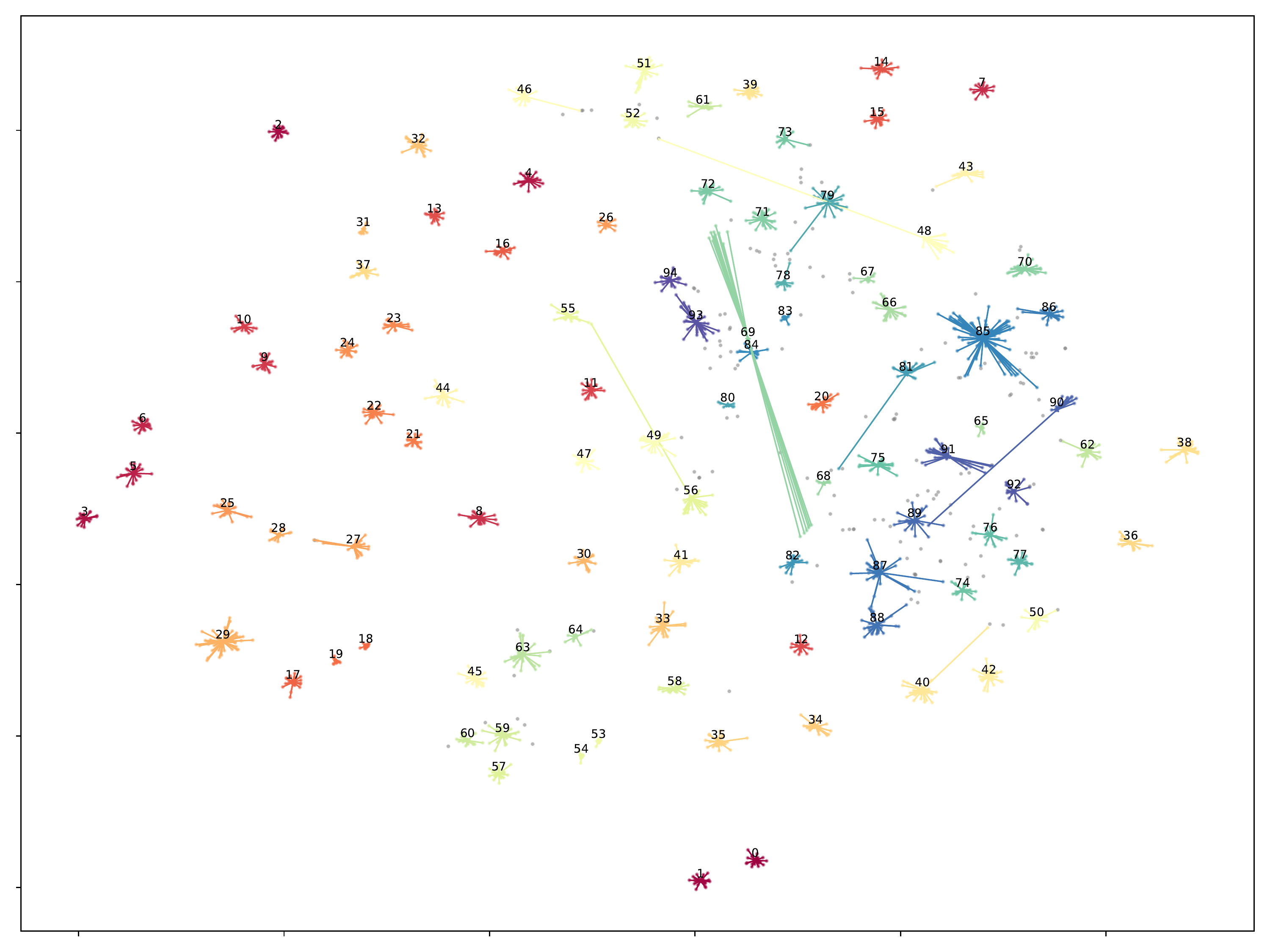}}
\subfigure[]{
\label{fig8d1}
\includegraphics[width=0.18\textwidth]{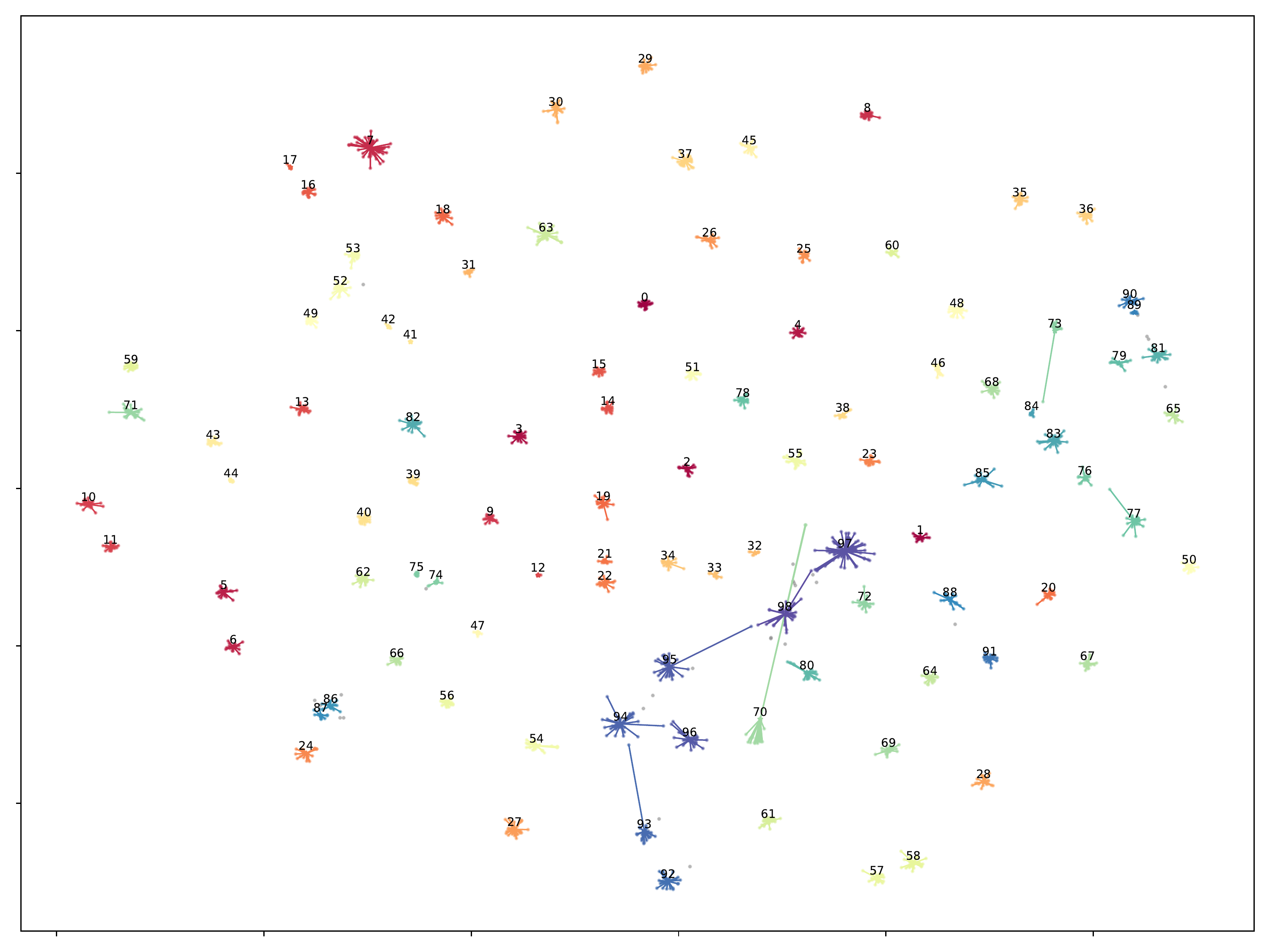}}
\subfigure[]{
\label{fig8e1}
\includegraphics[width=0.18\textwidth]{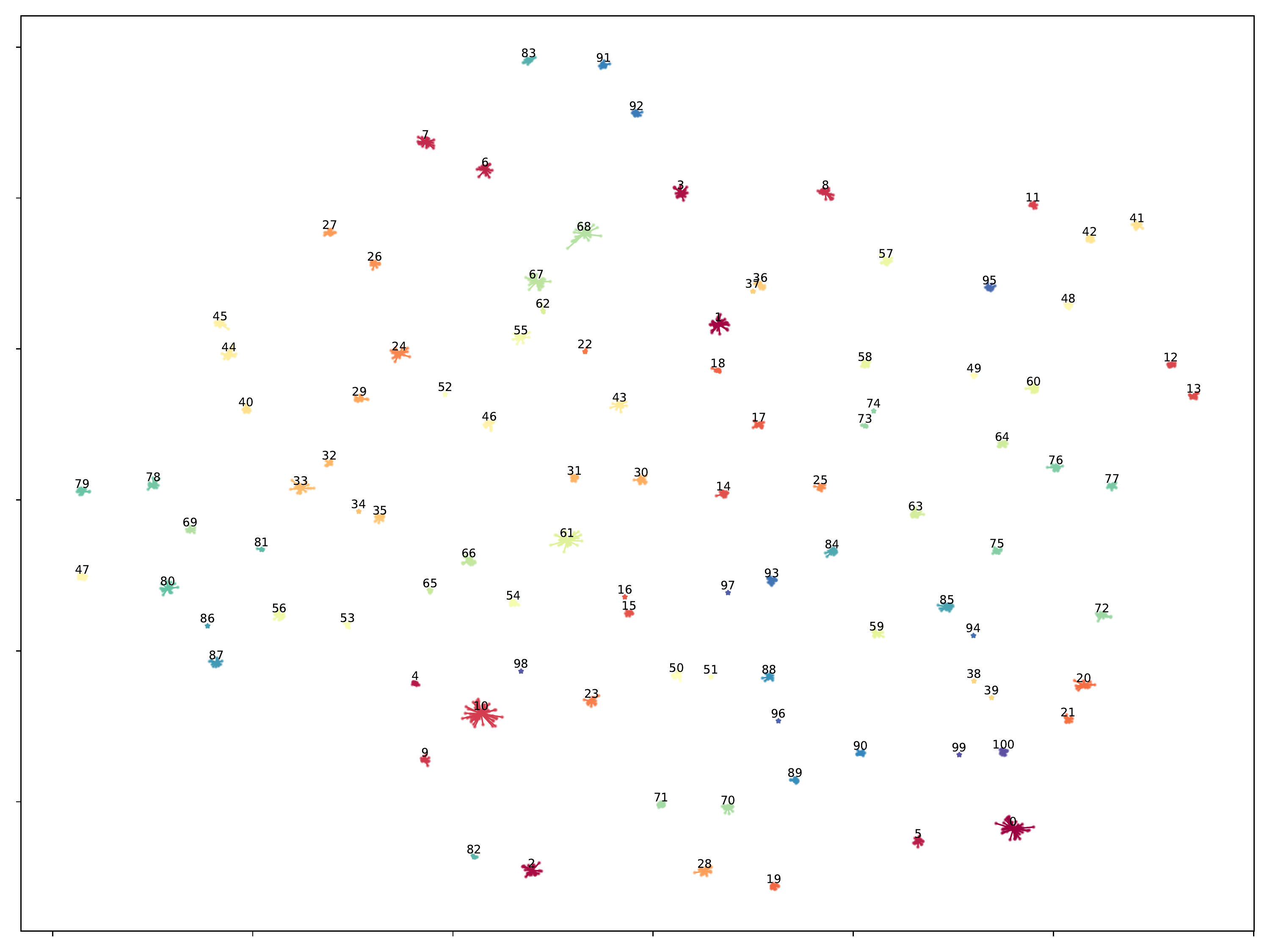}}
\caption{Distribution of features from CUB-200-2011 (first line) and Oxford-Flowers (second line). (a)-(e) and (f)-(j) are the clustering results of the 0th, 1st, 10th, 20th and 50th epochs, respectively.
\textbf{Best viewed on a monitor when zoomed in}.}
\label{fig8} 
\end{figure*}

To visualize the distribution of the extracted features, we embedded the features into a 2D plane by the $t$-SNE algorithm on CUB-200-2011 and Oxford-Flowers. In order to show the distribution of the intra-class features, we connect the intra-class features to their feature agent. The features are showed in Fig.~\ref{fig8}.


It can be found from Fig.~\ref{fig8} that at the beginning (epoch 0), CUB-200-2011 had poor initial distribution compared with Oxford-Flowers, and most of the examples are scattered on the feature space. after the 10-th epoch, the intra-class examples start to cluster, and finally move together at the 50-th epoch.

\section{Conclusion}
In this paper, we explored an unsupervised learning method of fine-grained image classification, and examined each module on the impact performance in the clustering learning process and discover three key modules that contribute to the performance of unsupervised FGVC. For feature extraction module, we found that a strong and powerful backbone can obtain better performance than weak backbone, the quality of the initial extracted feature distribution has an important impact on the clustering. It is easy to lead to the deterioration or failure especially when the initial feature distribution is poor. For the clustering module, HDBSCAN can produce stable clusters and achieve better performance than DBSCAN with fewer super-parameters. For the contrastive learning module, cluster-level contrastive loss with weighted feature agent can learn the parameters better. In each training epoch, the number of iterations is an important factor, and it should be set by balancing over-fitting and under-fitting. We hope these new observations and discussions can challenge some common beliefs and encourage people to rethink the importance of the three key modules in unsupervised FGVC.

Although the proposed method has achieved better results on multiple datasets, but it still needs further research. For example, the existing unsupervised contrastive learning can effectively learn the feature representation of a single example without clustering the whole training dataset. It is potentially learning method worthy of exploration.

\ifCLASSOPTIONcaptionsoff
  \newpage
\fi



\bibliographystyle{IEEEtran}
\bibliography{UFCL}

\vfill


\end{document}